\documentclass[final,5p,times,twocolumn,authoryear]{elsarticle}
\usepackage[T1]{fontenc}
\usepackage[utf8]{inputenc}
\usepackage{amsmath,amssymb,booktabs,multirow,array,tabularx}
\usepackage{xcolor,placeins}
\usepackage{etoolbox,xurl,flushend}
\usepackage[unicode,pdfversion=1.7]{hyperref}
\biboptions{authoryear,round}

\graphicspath{{figures/}}
\hypersetup{hidelinks,breaklinks=true,pdfcreator={Jialei.He},pdftitle={ContiLNN: Mitigating Slice Sampling Discontinuity with Liquid Neural Networks for Medical Image Restoration}}
\journal{Medical Image Analysis}
\begin{document}
\begin{frontmatter}
\title{ContiLNN: Mitigating Slice Sampling Discontinuity with Liquid Neural Networks for Medical Image Restoration}
\author[a,b]{Jialei He\fnref{equal}}
\author[a,b]{Enhe Liu\fnref{equal}}
\author[b]{Sifan Song}
\author[c]{Pengfei Jin}
\author[b]{Jionglong Su}
\author[a]{Hongbin Wang}
\author[b]{Zhixiang Lu}
\author[a,b]{Yanhao Huang}
\author[a,b]{Anteng Cai}
\author[b]{Zhengyong Jiang}
\author[a]{Jiaman Ding\corref{cor}}
\author[d]{S. Kevin Zhou\corref{cor}}
\author[a]{Jinfeng Wang\corref{cor}}
\ead{Jf.Jacob.Wang@gmail.com}
\cortext[cor]{Corresponding authors.}
\fntext[equal]{Jialei He and Enhe Liu contributed equally to this work.}
\affiliation[a]{organization={Faculty of Information Engineering and Automation, Kunming University of Science and Technology},city={Kunming},state={Yunnan},postcode={650504},country={China}}
\affiliation[b]{organization={Academy of Artificial Intelligence and Advanced Technology, Xi'an Jiaotong-Liverpool University},city={Suzhou},state={Jiangsu},postcode={215123},country={China}}
\affiliation[c]{organization={Department of Radiology, Massachusetts General Hospital and Harvard Medical School},city={Boston},state={Massachusetts},postcode={02115},country={United States}}
\affiliation[d]{organization={School of BME \& Suzhou Institute for Advanced Research, Center for Medical Imaging, Robotics, Analytic Computing \& Learning (MIRACLE), University of Science and Technology of China},city={Suzhou},state={Jiangsu},postcode={215000},country={China}}
\begin{abstract}
Anatomical continuity provides complementary information for medical image restoration, yet exploiting neighboring slices requires accounting for both local anatomy and variations in slice sampling. We introduce ContiLNN, which augments two-dimensional restoration networks with bidirectional closed-form continuous-time (Bi-CfC) modules to model cross-slice context through liquid neural dynamics while retaining in-plane feature extraction. Slice-index intervals modulate gates determined by local features and hidden states, allowing propagation to respond to sampling variations and anatomical context. Closed-form updates avoid iterative numerical integration. Alongside reconstruction supervision, cross-slice consistency aligns first- and second-order intensity differences with high-quality references to preserve anatomical variation. Distillation from a frozen two-dimensional backbone helps retain in-plane fidelity. Across five training seeds, ContiLNN improves mean peak signal-to-noise ratio (PSNR) by 0.1907, 1.0176 and 1.2482 dB over the two-dimensional baseline for CT denoising, MRI super-resolution and reduced-count PET restoration, respectively, while reducing root mean squared error in all three tasks. On the held-out CT patient, PSNR improves consistently, with a standard deviation of 0.0026 dB across seeds; gains are greater for MRI and PET. PET ablations support ordered propagation beyond additional pointwise capacity. Under contiguous training, Bi-CfC achieves higher fidelity than a bidirectional gated recurrent unit (Bi-GRU) with similar parameter counts and arithmetic costs across all tested sampling conditions, with lower measured latency and memory use. We also investigate the performance of both operators under sparse and irregular context using mixed-gap training. Experiments with fewer training patients and a second backbone further support data efficiency and backbone compatibility.
\end{abstract}
\end{frontmatter}
% Editor: Jialei.He. Keep natural paragraph spacing in short columns.
\raggedbottom
% Editor: Jialei.He
\section{Introduction}
Medical image restoration seeks to recover anatomical information from acquisitions constrained by radiation dose, scan duration or signal availability. Lower-dose CT reduces radiation exposure at the cost of increased noise \citep{ldctChallenge,ldctDataset}. In MRI, higher spatial resolution competes with acquisition time and signal-to-noise ratio, motivating super-resolution methods that recover structural detail from lower-resolution observations \citep{beyondNyquistMRI}. In PET, lower administered tracer activity can reduce radiation exposure, and shorter acquisitions can reduce scan burden; both reduce detected counts, increasing noise and making local uptake differences harder to distinguish \citep{spachTransformer,arganPET}. These tasks estimate high-quality (HQ) images from low-quality (LQ) inputs. Advances in convolutional, Transformer and state-space networks have substantially improved in-plane recovery, providing efficient backbones for a range of medical restoration tasks \citep{restormer,restoreRWKV,dasmamba,taskadaptiveRoutingMIR,ldvqMIR,tatMIR}.

\begin{figure*}[!t]
\centering
\includegraphics[width=0.98\textwidth]{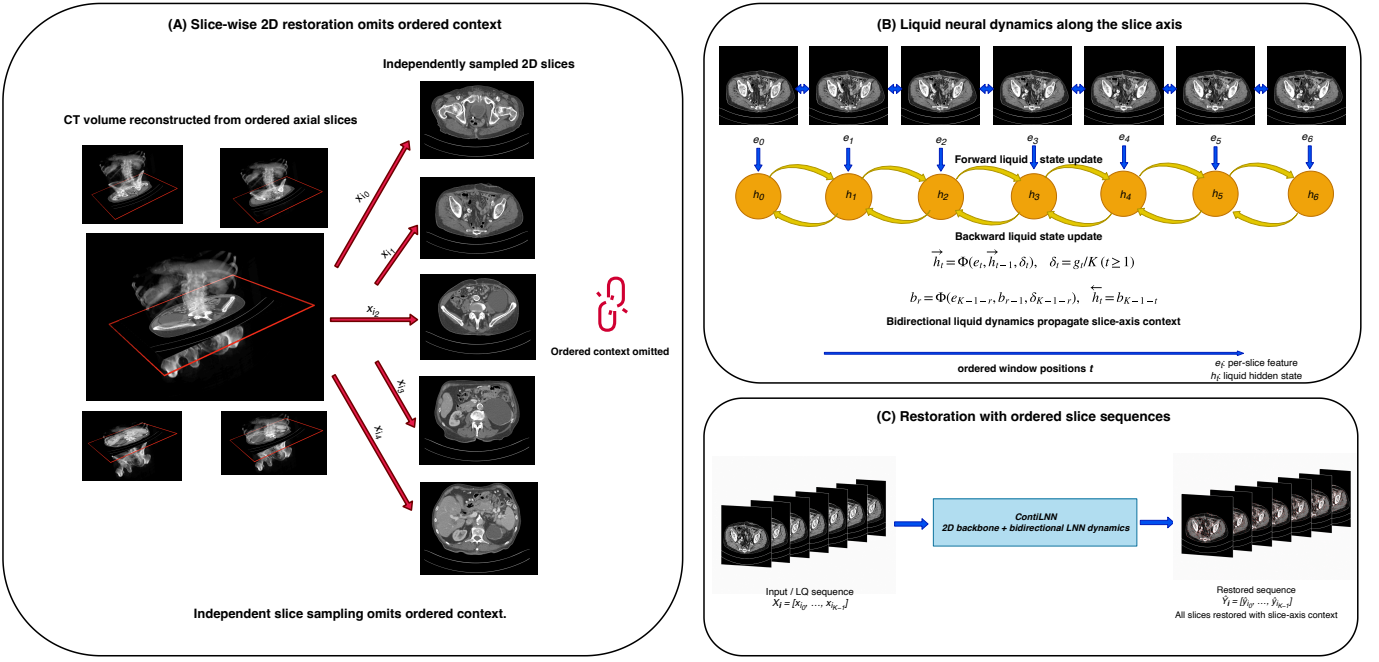}
\caption{Ordered-slice restoration with ContiLNN. (A) Independent 2D processing. (B) Bidirectional liquid propagation. (C) One restored output per observed slice. Original slice indices $i_t$ identify the $K$ inputs $x_{i_t}$ and outputs $\widehat y_{i_t}$; $t$ denotes window position. The direction-specific closed-form continuous-time (CfC) update $\Phi$ propagates states $h_t$ using features $e_t$ and normalized intervals $\delta_t$ derived from gaps $g_t$. The backward traversal state $b_r$ is realigned to the original order; interval conventions and alignment are defined in Eqs.~\ref{eq:interval_input} and \ref{eq:bidirectional}.}
\label{fig:methodology}
\end{figure*}

The anatomical evidence in a volume extends beyond any individual slice. Adjacent images sample structures that evolve along the slice axis, yet independent two-dimensional (2D) processing does not explicitly retain this ordered relationship (Fig.~\ref{fig:methodology}A). The imbalance between in-plane and through-plane information is a recognized challenge in volumetric medical image modeling \citep{mnetAnisotropic}.

A relevant clinical application arises when only a subset of slices from an existing image series is available for retrospective restoration. The available slices may be regularly or irregularly spaced in the original index sequence, so adjacent inputs need not correspond to consecutive slices in the source series. Their original indices preserve this distinction. This setting motivates cross-slice feature propagation that uses neighboring anatomy while accounting for the index gaps between available observations, with a restored output retained at each supplied position.

% Editor: Jialei.He. Keep this paragraph opening with its continuation.
\begingroup
\clubpenalty=10000
Local 2.5D modeling combines in-plane feature extraction with information from neighboring slices. Three-dimensional (3D) restoration networks instead model context through volumetric feature maps, as in convolutional models for MRI super-resolution \citep{beyondNyquistMRI} and Spach Transformer for PET denoising \citep{spachTransformer}. Storage for these feature maps scales with their spatial extent and channel width. Neighboring CT slices have also been incorporated into convolutional encoder--decoders \citep{cpce3d}. Adjacent-slice supervision and multi-view inference provide further ways to use through-plane information \citep{noise2context,twoPointFiveDPETDiffusion}. These approaches establish several roles for local anatomical context. ContiLNN represents this context as an evolving sequence of feature states, with an explicit slice-index interval and a restored output at every supplied position.
\par
\endgroup

Liquid neural networks provide a continuous-time model of input-dependent state evolution \citep{ltc}. Closed-form continuous-time networks (CfCs) approximate this evolution through interval-dependent updates without repeated numerical integration \citep{cfc}. Their explicit interval input is particularly relevant when the separation between observations varies. For medical image restoration, we propagate states in slice order and condition each update on a normalized slice-index interval. Forward and backward states then communicate information between preceding and following observations (Fig.~\ref{fig:methodology}B). This links the anatomical ordering of the input to the dynamics used to refine its features.

Here we introduce ContiLNN, which augments 2D restoration backbones with bidirectional liquid dynamics along the slice axis. Each local sequence retains the correspondence between its input slices and restored outputs (Fig.~\ref{fig:methodology}C). Bidirectional CfC (Bi-CfC) modules propagate context at selected levels of the feature hierarchy, while two complementary regularizers guide the resulting images: cross-slice consistency preserves first- and second-order intensity changes in the HQ sequence, and distillation from a frozen 2D backbone retains agreement with its in-plane estimates. We develop this design on Restore-RWKV \citep{restoreRWKV} and apply it to DASMamba \citep{dasmamba} to examine compatibility with a second backbone.

Our contributions are threefold:
\begin{enumerate}
\item \textbf{Liquid dynamics for medical image restoration.} We introduce ContiLNN to propagate anatomical information across ordered slices through bidirectional liquid dynamics. Slice-index intervals condition the closed-form updates, allowing state propagation to respond to sampling changes.
\item \textbf{Augmentation of 2D restoration backbones.} Bi-CfC modules refine selected feature pathways while retaining the backbone's in-plane blocks. Cross-slice intensity consistency and 2D backbone distillation jointly preserve reference anatomical variation and in-plane fidelity.
\item \textbf{Evaluation across modalities and sampling settings.} CT, MRI and PET experiments assess restoration quality. Case-level analyses, comparisons of two operators under regular, sparse and irregular sampling, limited-patient training, backbone studies and ablations assess sampling adaptation, data efficiency and backbone compatibility.
\end{enumerate}
% Editor: Jialei.He
\section{Related work}
\subsection{Medical image restoration}
Learning-based restoration has progressed from local residual prediction to efficient modeling of long-range image structure. Convolutional restoration networks learn local image priors from paired images or neighboring observations \citep{cpce3d,noise2context}. Transformer models extend the spatial reach of these representations through attention, improving the recovery of fine structures and long-range dependencies \citep{swinir,restormer}. RWKV combines recurrent inference with parallelizable training, and its visual adaptations provide another route to efficient spatial modeling \citep{rwkv,visionRWKV}. MambaIR, DenoMamba, CT-Mamba, Restore-RWKV and DASMamba apply state-space or recurrent-style representations to natural and medical image restoration \citep{mamba,mambair,denomamba,ctmamba,restoreRWKV,dasmamba}. These methods supply the in-plane feature hierarchies on which ContiLNN builds. Whereas our previous work FedFTN \citep{zhou2023fedftn} addresses institutional heterogeneity through personalized federated learning for low-count PET denoising, ContiLNN models cross-slice anatomical context through interval-conditioned bidirectional liquid dynamics for CT, MRI, and PET restoration.
\subsection{Slice-axis context}
The choice of spatial representation determines how neighboring slices contribute to a prediction. Slice-wise networks are compatible with high-resolution images, whereas volumetric networks directly couple in-plane and through-plane structure. Anisotropic sampling complicates this balance because a fixed number of neighboring slices can span substantially different anatomical extents \citep{mnetAnisotropic}. Consequently, slice count alone does not describe the context available to a model; the sampling intervals also matter.

Restoration methods use through-plane information in several distinct ways. Hybrid 2D/3D convolution can aggregate neighboring CT slices for center-slice denoising \citep{cpce3d}, while volumetric MRI networks learn features across all three spatial axes for super-resolution \citep{beyondNyquistMRI}. Other approaches use adjacent slices as training supervision or combine predictions from multiple views \citep{noise2context,twoPointFiveDPETDiffusion}. Slice-based reconstruction addresses a related problem by recovering volumes or anatomical surfaces from sparse observations \citep{nesvor,slice2mesh}. ContiLNN maintains an ordered feature sequence and restores every supplied slice. Its interval inputs are derived from the original index gaps between selected observations, allowing feature updates to respond to both local image content and the sampling of neighboring context.

\subsection{Liquid neural dynamics}
Continuous-time models describe how a hidden representation evolves between observations. Neural ordinary differential equations (ODEs) express this evolution through a learned vector field \citep{neuralODE}, while latent ODE models provide continuous-time latent representations for irregularly observed sequences \citep{latentODE}. Liquid time-constant networks introduce input-dependent dynamics, allowing the effective transition to vary with the observation \citep{ltc}. CfCs approximate these dynamics with closed-form updates that retain an explicit dependence on the supplied interval \citep{cfc}.

This dependence offers a useful connection between continuous-time modeling and slice sampling. An update can respond to a larger index gap without treating it as the same transition as two consecutive slices. ContiLNN applies this principle to dense features at corresponding in-plane locations, with two directions of propagation supplying complementary anatomical context. We evaluate Bi-CfC across sparse and irregular neighboring contexts after contiguous training, and investigate mixed-gap training separately as a way to broaden exposure to sampling variations.

\subsection{Consistency and distillation}
Consistency regularization encourages agreement under input or model perturbations \citep{meanTeacher,temporalEnsembling}, while knowledge distillation transfers information through predictions from a reference model \citep{hintonKD}. In restoration, these constraints complement direct fidelity to the reference image. For an ordered slice sequence, neighboring reference images need not have identical intensities: anatomical boundaries and uptake patterns can change substantially with position. We therefore match the first- and second-order differences of restored and reference sequences. A separate 2D backbone distillation term encourages agreement with predictions from a frozen, pretrained 2D backbone. These two forms of regularization preserve useful in-plane estimates while learning the additional information supplied by neighboring slices.

% Editor: Jialei.He
\section{Method}
\subsection{Ordered-slice restoration}
ContiLNN restores a local sequence of observed slices while preserving their positional correspondence. Let $X=\{x_i\}_{i=1}^{N}$ denote an LQ volume with $N$ ordered slices and $Y=\{y_i\}_{i=1}^{N}$ its HQ reference. A local window contains $K\leq T$ slices, where $T$ is the maximum window length. Its increasing original indices are $\boldsymbol{i}=(i_0,\ldots,i_{K-1})$, and its paired input and reference sequences are
\begin{equation}
\begin{aligned}
X_{\boldsymbol{i}}&=[x_{i_0},\ldots,x_{i_{K-1}}],\\
Y_{\boldsymbol{i}}&=[y_{i_0},\ldots,y_{i_{K-1}}].
\end{aligned}
\label{eq:sequences}
\end{equation}
The model learns $\mathcal F_{\theta}:X_{\boldsymbol{i}}\mapsto\widehat Y_{\boldsymbol{i}}$, where $\theta$ denotes its trainable parameters and $\widehat Y_{\boldsymbol{i}}=[\widehat y_{i_0},\ldots,\widehat y_{i_{K-1}}]$ contains one restored image for each input position. Every output receives direct supervision from the corresponding HQ slice.

The slice-index gap specifies the sampling interval between consecutive observations in the window:
\begin{equation}
g_t=i_t-i_{t-1},\qquad t=1,\ldots,K-1.
\label{eq:gap}
\end{equation}
Fixed gaps of one, two and three select consecutive slices, every second slice and every third slice, respectively. Indices are retained from the source series before slice selection rather than reassigned consecutively afterwards. For example, available slices with indices $[1,3,5,7]$ have successive gaps $(2,2,2)$, whereas indices $[1,2,3,5,6]$ give gaps $(1,1,2,1)$. These gaps describe the separation between available observations in the original index sequence. ContiLNN restores each supplied slice at its corresponding position.

\begin{figure*}[!t]
\centering
\includegraphics[width=0.98\textwidth]{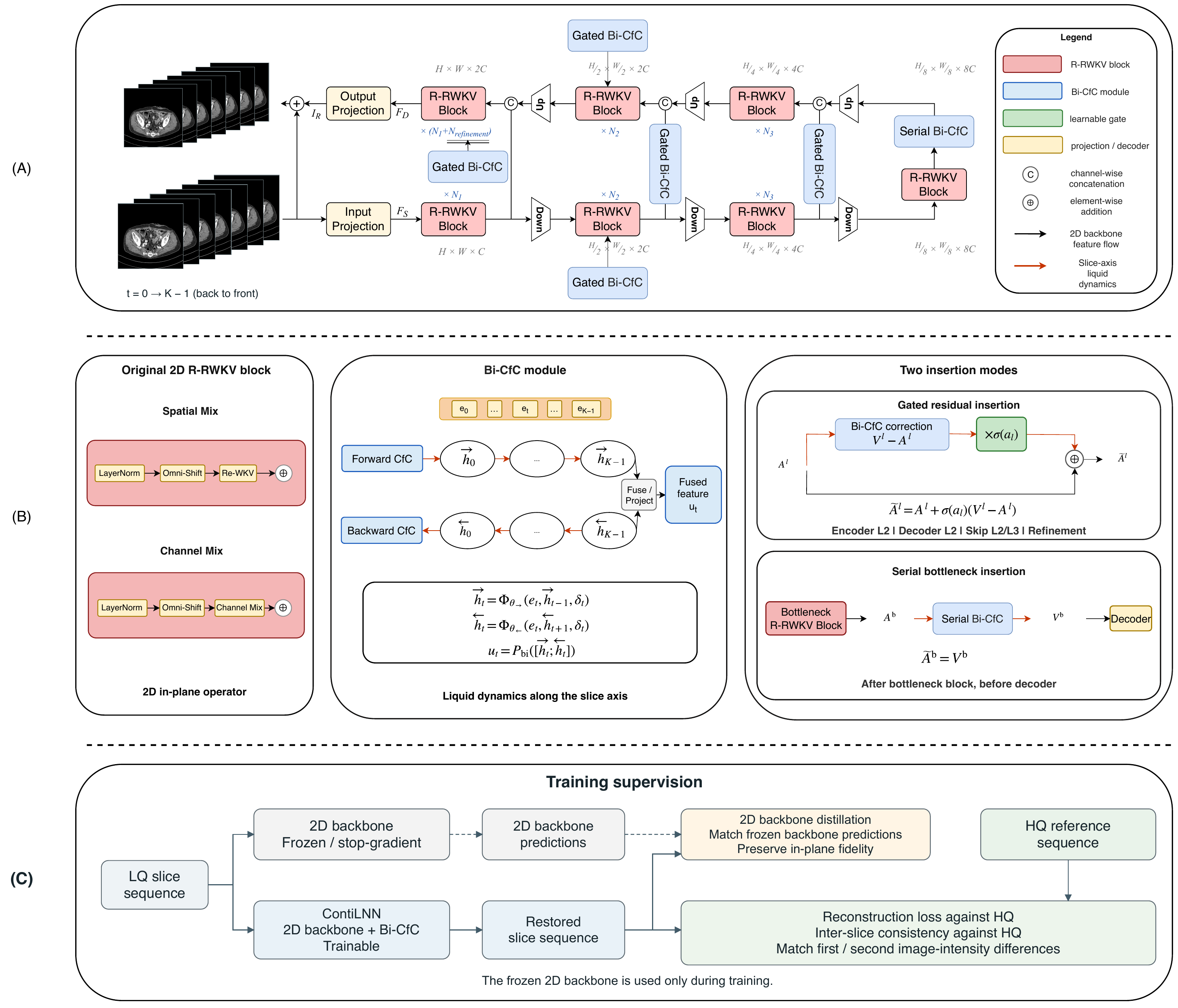}
\caption{ContiLNN architecture based on Restore-RWKV \citep{restoreRWKV}. (A) Bi-CfC modules augment selected feature pathways; the paired CT sequences show the same seven observed positions, ordered back to front. (B) Backbone blocks, the bidirectional liquid operator, and gated residual or serial insertion. Reverse traversal preserves feature--interval pairing (Eq.~\ref{eq:bidirectional}). (C) Training supervision. High-quality references supervise reconstruction and first- and second-order inter-slice intensity differences. A frozen 2D backbone supplies single-slice predictions for distillation during training; the ContiLNN backbone and Bi-CfC modules remain trainable.}
\label{fig:architecture}
\end{figure*}

\subsection{Bidirectional liquid dynamics}
The bidirectional closed-form continuous-time module, denoted Bi-CfC, is the slice-axis operator of ContiLNN (Fig.~\ref{fig:architecture}B). Its hidden state propagates over ordered slice features, with an explicit interval input associated with each observation. The continuous-time formulation motivating this operator represents state evolution as
\begin{equation}
\frac{\mathrm d h^l(q)}{\mathrm d q}
=G_{\theta}^l\bigl(h^l(q),e^l(q)\bigr),
\label{eq:dynamics}
\end{equation}
where $q$ is a continuous evolution coordinate, $l$ indexes a feature level, $h^l$ is its hidden state, $e^l$ is the input feature and $G_{\theta}^l$ is a learned vector field. CfC develops interval-dependent neural transitions from an approximate closed-form treatment of liquid dynamics \citep{cfc}. Based on this approximation, we use the default gated CfC parameterization to evaluate the state directly from the current feature, incoming state and supplied interval, without a numerical ODE solver.

We distinguish position within a window from the interval supplied to the dynamics. The index gaps are normalized by the actual number of observations $K$ to obtain the interval inputs defined below. For $K>1$, the within-window position and interval input are
\begin{equation}
p_t=\frac{t}{K-1},\qquad
\delta_t=
\begin{cases}
g_1/K,&t=0,\\
g_t/K,&t=1,\ldots,K-1.
\end{cases}
\label{eq:interval_input}
\end{equation}
The normalization uses the actual number of observations $K$, including short windows. Under fixed-gap sampling, every observation receives $g/K$; contiguous input gives $1/K$. The position $p_t$ specifies relative order, whereas $\delta_t$ supplies the slice-index interval attached to the current observation. Both branches preserve this feature--interval pairing. The single-slice boundary convention is specified in Supplementary Section~A.

At each in-plane location, let $f_t^l\in\mathbb R^{C_l}$ be the channel vector of slice $i_t$, where $C_l$ is the feature width. A learned affine position projection $P_{\mathrm{pos}}^l$ gives
\begin{equation}
e_t^l=f_t^l+P_{\mathrm{pos}}^l(p_t).
\label{eq:embedding}
\end{equation}
The embedded feature $e_t^l$ retains the original observation's within-window position in both traversal directions. The resulting feature and interval are supplied together to the directional cell.

The cell first forms a joint representation of the current feature and incoming state. For one feature level and direction, suppress the superscripts and write $C=C_l$ and $M=M_l$, where $M_l$ is the width of the cell's internal feature network. With $e_t,h_{t-1}\in\mathbb R^C$, the joint representation and candidate states are
\begin{equation}
\begin{aligned}
z_t&=\rho\!\left(W_z[e_t;h_{t-1}]+v_z\right),\\
c_t^{(j)}&=\tanh(W_jz_t+v_j),\qquad j\in\{1,2\}.
\end{aligned}
\label{eq:cfc_candidates}
\end{equation}
Here $[\,\cdot\,;\,\cdot\,]$ denotes channel concatenation, $W_z\in\mathbb R^{M\times2C}$ and $v_z\in\mathbb R^M$ are learned parameters, and $\rho$ is the scaled hyperbolic-tangent activation specified in Supplementary Section~A. Thus, $z_t\in\mathbb R^M$ combines present image content with the recurrent context. Each candidate head has weights $W_j\in\mathbb R^{C\times M}$ and bias $v_j\in\mathbb R^C$, producing $c_t^{(j)}\in(-1,1)^C$. The internal width $M$ is distinct from the recurrent-state width $C$.

Two further affine heads determine how the supplied interval modulates the mixture:
\begin{equation}
\begin{aligned}
\eta_t&=Q_a(z_t)=W_a z_t+v_a,\\
\xi_t&=Q_b(z_t)=W_b z_t+v_b,\\
\alpha_t&=\sigma(\delta_t\eta_t+\xi_t).
\end{aligned}
\label{eq:cfc_gate}
\end{equation}
The matrices $W_a,W_b\in\mathbb R^{C\times M}$ and biases $v_a,v_b\in\mathbb R^C$ are learned independently of the candidate heads. The vectors $\eta_t$ and $\xi_t$ are the interval coefficient and intercept, respectively; the scalar $\delta_t$ is broadcast across channels. The elementwise sigmoid $\sigma$ yields $\alpha_t\in(0,1)^C$. Because the affine coefficient $\eta_t$ can take either sign, different feature channels can respond differently to a larger interval.

The updated state is a channelwise convex combination of the two candidates:
\begin{equation}
\begin{aligned}
h_t&=(1-\alpha_t)\odot c_t^{(1)}
      +\alpha_t\odot c_t^{(2)}\\
&=c_t^{(1)}+\alpha_t\odot
       \bigl(c_t^{(2)}-c_t^{(1)}\bigr).
\end{aligned}
\label{eq:cfc_update}
\end{equation}
Here $\odot$ denotes the elementwise product. Each state channel lies between its two candidate values, keeping the directional hidden state in $(-1,1)^C$ before the subsequent affine feature projection. We denote the complete transition in Eqs.~\ref{eq:cfc_candidates}--\ref{eq:cfc_update} by $\Phi_{\theta}(e_t,h_{t-1},\delta_t)$. Both candidates depend on $h_{t-1}$, allowing the update to retain information from previously visited observations.

The gate provides an explicit, content-dependent response to sampling. Holding the incoming feature $e_t$ and state $h_{t-1}$ fixed gives
\begin{equation}
\begin{aligned}
\frac{\partial h_t}{\partial\delta_t}
&=\alpha_t\odot(1-\alpha_t)\odot\eta_t
 \odot\bigl(c_t^{(2)}-c_t^{(1)}\bigr),\\
\left|\frac{\partial h_t}{\partial\delta_t}\right|
&\leq\frac14|\eta_t|\odot
 \bigl|c_t^{(2)}-c_t^{(1)}\bigr|.
\end{aligned}
\label{eq:interval_response}
\end{equation}
Absolute values and the inequality are elementwise. The derivative isolates the direct response to the interval from changes carried through the recurrent state: its sign and magnitude depend on the learned coefficient, gate and candidate separation. The model therefore adapts its state update to the supplied sampling interval and local features while retaining algebraic evaluation of each transition.

Independent forward and backward cells apply this transition to the ordered sequence. Let $r=0,\ldots,K-1$ index backward traversal, and let $b_r^l$ be its recurrent state. The recurrences and alignment to the original positions are
\begin{equation}
\begin{aligned}
\overrightarrow h_t^l
&=\Phi_{\theta_{\rightarrow}}^l
  (e_t^l,\overrightarrow h_{t-1}^l,\delta_t),\\
b_r^l
&=\Phi_{\theta_{\leftarrow}}^l
  (e_{K-1-r}^l,b_{r-1}^l,\delta_{K-1-r}),\\
\overleftarrow h_t^l&=b_{K-1-t}^l.
\end{aligned}
\label{eq:bidirectional}
\end{equation}
Here $\theta_{\rightarrow}$ and $\theta_{\leftarrow}$ are separate parameter sets for the same cell architecture, with initial states $\overrightarrow h_{-1}^l=b_{-1}^l=0$. Reversing the feature--interval pairs preserves $e_t^l$ and $\delta_t$ at original position $t$. For irregular inputs, $\delta_t$ is a sampling descriptor associated with observation $t$; the index separation to position $t+1$ would instead correspond to $\delta_{t+1}$. The implemented backward branch retains the observation-specific descriptor. An example of this pairing is provided in Supplementary Section~A.

An affine projection fuses the two states aligned at each original position:
\begin{equation}
\begin{aligned}
u_t^l&=P_{\mathrm{bi}}^l
\bigl([\overrightarrow h_t^l;\overleftarrow h_t^l]\bigr)\\
&=W_{\mathrm{bi}}^l
[\overrightarrow h_t^l;\overleftarrow h_t^l]+v_{\mathrm{bi}}^l.
\end{aligned}
\label{eq:fusion}
\end{equation}
The learned weights $W_{\mathrm{bi}}^l\in\mathbb R^{C_l\times2C_l}$ and bias $v_{\mathrm{bi}}^l\in\mathbb R^{C_l}$ return a refined vector $u_t^l\in\mathbb R^{C_l}$. Each output thus combines preceding and following observations at the same slice position before being returned to the backbone's feature layout.

\subsection{Backbone augmentation}
Bi-CfC acts on slice sequences extracted from an otherwise two-dimensional feature hierarchy. The R-RWKV block is the original in-plane building block of Restore-RWKV: its omnidirectional token shift (Omni-Shift) aggregates local spatial information, and its recurrent WKV attention (Re-WKV) models longer-range dependencies across in-plane scan directions \citep{restoreRWKV}. ContiLNN retains these operations and adds Bi-CfC at selected feature pathways (Fig.~\ref{fig:architecture}A,B). For a batch of $B$ windows, let $A^l$ be the backbone feature tensor with spatial dimensions $H_l\times W_l$. A rearrangement $\mathcal R_l$ exposes the ordered channel vectors at each spatial location:
\begin{equation}
\begin{gathered}
A^l\in\mathbb R^{BK\times C_l\times H_l\times W_l}
\xrightarrow{\ \mathcal R_l\ }
F^l\in\mathbb R^{BH_lW_l\times K\times C_l},\\
U^l=\Psi_l(F^l,\boldsymbol\delta),\qquad
V^l=\mathcal R_l^{-1}(U^l).
\end{gathered}
\label{eq:reshape}
\end{equation}
Here $U^l$ is the processed feature sequence and $V^l$ its restored 2D layout. The rearrangement separates the batch and slice dimensions before grouping the batch and spatial locations. The operator $\Psi_l$ applies the position embedding, bidirectional updates and fusion in Eqs.~\ref{eq:embedding}--\ref{eq:fusion}; $\boldsymbol\delta$ contains the $K$ interval inputs. The same operator is shared across in-plane locations, and $\mathcal R_l^{-1}$ restores the original tensor layout.

The two insertion modes have complementary roles. Serial insertion introduces sequence information at the bottleneck before decoding; gated residual insertion learns the correction strength along the remaining selected pathways (Fig.~\ref{fig:architecture}B):
\begin{equation}
\widetilde A^l=
\begin{cases}
V^l, & l\in\mathcal S_{\mathrm b},\\
A^l+\sigma(a_l)(V^l-A^l),&l\in\mathcal S_{\mathrm g},\\
A^l,&l\notin\mathcal S_{\mathrm b}\cup\mathcal S_{\mathrm g}.
\end{cases}
\label{eq:insertion}
\end{equation}
Here $\mathcal S_{\mathrm b}$ and $\mathcal S_{\mathrm g}$ identify the bottleneck and gated pathways, $a_l$ is a learnable scalar gate logit, and $\sigma$ is the sigmoid function. Subsequent backbone blocks process $\widetilde A^l$ in their original 2D layout.

Restore-RWKV supplies the principal backbone, with Bi-CfC augmenting its intermediate encoder and decoder blocks, selected skips, bottleneck and late refinement pathway. The same insertion roles are applied to DASMamba to examine backbone compatibility. Supplementary Section~A reports the representative Restore-RWKV topology and module settings. This design enables slice-axis modeling within existing feature hierarchies while keeping sequence propagation local to one patient or scan.

The selected pathways address different stages of restoration. Intermediate encoder and decoder features provide reduced-resolution representations that retain local structure for cross-slice exchange. Skip-path modules carry this context from encoding to decoding, while serial bottleneck processing makes the deepest representation sequence-aware before upsampling. The late refinement pathway supplies a final contextual correction. Gates regulate the contribution of the transformed sequence features along the remaining pathways, allowing their strength to be learned jointly with the in-plane backbone.

\subsection{Training objective}
The training objective combines reconstruction, cross-slice consistency and 2D backbone distillation (Fig.~\ref{fig:architecture}C). ContiLNN restores each input sequence, while a frozen copy of the pretrained 2D backbone processes the same slices independently to provide distillation targets. Paired HQ images supervise reconstruction and cross-slice intensity differences; the frozen backbone predictions guide preservation of in-plane fidelity. Gradients update the ContiLNN backbone and Bi-CfC modules, but do not pass through the frozen backbone, which is used only during training. The resulting objective is
\begin{equation}
\mathcal L=\mathcal L_{\mathrm{rec}}
+\lambda_{\mathrm{cons}}\mathcal L_{\mathrm{cons}}
+\lambda_{\mathrm{KD}}\mathcal L_{\mathrm{KD}},
\label{eq:total_loss}
\end{equation}
where $\lambda_{\mathrm{cons}}$ and $\lambda_{\mathrm{KD}}$ weight cross-slice consistency and 2D backbone distillation. For the Restore-RWKV backbone,
\begin{equation}
\begin{aligned}
\mathcal L_{\mathrm{rec}}
&=\frac1K\sum_{t=0}^{K-1}
\|\widehat y_{i_t}-y_{i_t}\|_1,\\
\mathcal L_{\mathrm{KD}}
&=\frac1K\sum_{t=0}^{K-1}
\|\widehat y_{i_t}-\overline y_{i_t}\|_1,
\end{aligned}
\label{eq:fidelity_distillation}
\end{equation}
where $\|\cdot\|_1$ denotes mean absolute error over image pixels and $\overline y_{i_t}$ is the prediction of the frozen 2D backbone for $x_{i_t}$. The ContiLNN backbone and Bi-CfC parameters remain trainable. For DASMamba, its native reconstruction penalties are retained within $\mathcal L_{\mathrm{rec}}$, while the remaining workflow follows ContiLNN-RWKV.

Cross-slice consistency follows anatomical change in the reference sequence. Define the first and second finite differences as
\begin{equation}
\begin{aligned}
D_1y_t&=y_{i_{t+1}}-y_{i_t},\\
D_2y_t&=y_{i_{t+1}}-2y_{i_t}+y_{i_{t-1}},
\end{aligned}
\label{eq:differences}
\end{equation}
with identical definitions for predictions. The consistency objective is
\begin{equation}
\begin{aligned}
\mathcal L_{\mathrm{cons}}&=
\mathcal L_{\mathrm{cons}}^{(1)}
+\beta_2\mathcal L_{\mathrm{cons}}^{(2)},\\
\mathcal L_{\mathrm{cons}}^{(1)}&=
\frac1{K-1}\sum_{t=0}^{K-2}
\|D_1\widehat y_t-D_1y_t\|_1,\\
\mathcal L_{\mathrm{cons}}^{(2)}&=
\frac1{K-2}\sum_{t=1}^{K-2}
\|D_2\widehat y_t-D_2y_t\|_1,
\end{aligned}
\label{eq:consistency}
\end{equation}
where $\beta_2$ weights second-order changes. The corresponding term is zero when a short window contains too few slices to evaluate its difference. These are finite differences over the observed sequence, including any index gaps. The first-order term matches the intensity change between successive observations; the second-order term matches how this change varies across the local sequence. Equivalently, defining the restoration error as $r_t=\widehat y_{i_t}-y_{i_t}$ gives
\begin{equation}
\begin{aligned}
D_1\widehat y_t-D_1y_t&=D_1r_t,\\
D_2\widehat y_t-D_2y_t&=D_2r_t.
\end{aligned}
\label{eq:error_consistency}
\end{equation}
The penalties thus constrain inter-slice variation in restoration error while retaining the anatomical changes encoded by the reference. Reconstruction anchors the absolute intensity of each output, and distillation retains agreement with the pretrained in-plane estimate. Numerical weights are listed in Supplementary Section~A.

\subsection{Inference}
Restoration covers the full set of observed slices through local, case-specific windows. Training retains short terminal windows, and inference combines predictions from overlapping window tilings. For slice $i$, let $\mathcal W(i)$ be the set of windows that contain it and $\widehat y_i^{(w)}$ the prediction from window $w$. The fused output is
\begin{equation}
\widehat y_i=\frac1{|\mathcal W(i)|}
\sum_{w\in\mathcal W(i)}\widehat y_i^{(w)}.
\label{eq:inference}
\end{equation}
Here $|\mathcal W(i)|$ is the number of windows covering slice $i$. Outputs are returned to their original slice order. The tilings retain every eligible real slice, and each hidden state is initialized within its own window. The window length and tiling offsets are specified in Supplementary Section~A. Sparse and irregular context use the same position-aligned output principle, allowing the sampling experiments to change contextual support while retaining the evaluated images.

% Editor: Jialei.He
% Editor: Jialei.He
\begin{table*}[!t]
\centering
\caption{Restoration results for MRI super-resolution, CT denoising and reduced-count PET restoration. Published reference results in the upper and lower blocks follow the Restore-RWKV and DASMamba studies, respectively \citep{restoreRWKV,dasmamba}; the upper published block uses joint training across the three modalities. In the lower published block, $^*$ marks methods re-implemented by the DASMamba authors. Each block ends with the corresponding backbone reproduced separately for each modality and its ContiLNN counterpart. The reproduced backbones use single-slice (2D) input, whereas ContiLNN uses local ordered-slice (2.5D) input. ContiLNN values are mean$\pm$sample standard deviation across five training seeds. Bold values indicate improvements over the corresponding reproduced backbone within each block. PSNR is reported in dB.}
\label{tab:main_results}
\scriptsize
\setlength{\tabcolsep}{2.2pt}
\renewcommand{\arraystretch}{1.12}
\resizebox{\textwidth}{!}{%
\begin{tabular}{lccccccccc}
\toprule
\multirow{2}{*}{Method} & \multicolumn{3}{c}{MRI super-resolution} & \multicolumn{3}{c}{CT denoising} & \multicolumn{3}{c}{PET restoration} \\
\cmidrule(lr){2-4}\cmidrule(lr){5-7}\cmidrule(lr){8-10}
 & PSNR $\uparrow$ & SSIM $\uparrow$ & RMSE $\downarrow$ & PSNR $\uparrow$ & SSIM $\uparrow$ & RMSE $\downarrow$ & PSNR $\uparrow$ & SSIM $\uparrow$ & RMSE $\downarrow$ \\
\midrule
\multicolumn{10}{l}{\textit{Published reference results: Restore-RWKV study}} \\
\addlinespace[2pt]
ARGAN \citep{arganPET} & 30.0800 & 0.9083 & 35.7999 & 32.9200 & 0.9111 & 9.2110 & 36.7500 & 0.9389 & 0.0907 \\
DenoMamba \citep{denomamba} & 30.3200 & 0.9091 & 34.6972 & 33.1800 & 0.9115 & 8.9512 & 36.8100 & 0.9367 & 0.0895 \\
MambaIR \citep{mambair} & 31.3100 & 0.9305 & 31.3150 & 33.5000 & 0.9165 & 8.6345 & 37.1700 & 0.9458 & 0.0864 \\
AirNet \citep{airnet} & 31.3900 & 0.9316 & 31.1141 & 33.6200 & 0.9176 & 8.5226 & 37.1700 & 0.9451 & 0.0864 \\
DRMC \citep{drmc} & 29.5500 & 0.9032 & 38.1691 & 33.2800 & 0.9153 & 8.8674 & 36.1900 & 0.9376 & 0.0960 \\
AMIR \citep{taskadaptiveRoutingMIR} & 32.0300 & 0.9396 & 29.0988 & 33.7000 & 0.9182 & 8.4520 & 37.1200 & 0.9475 & 0.0876 \\
Restore-RWKV routing \citep{restoreRWKV} & 32.1000 & 0.9408 & 28.9507 & 33.7400 & 0.9189 & 8.4148 & 37.3200 & 0.9479 & 0.0852 \\
\midrule
Restore-RWKV (reproduced) \citep{restoreRWKV} & 32.0650 & 0.9403 & 29.0337 & 33.7642 & 0.9189 & 8.3950 & 37.2722 & 0.9474 & 0.0860 \\
ContiLNN-RWKV (ours) & \textbf{33.0826$\pm$0.0261} & \textbf{0.9485$\pm$0.0003} & \textbf{25.9882$\pm$0.0809} & \textbf{33.9549$\pm$0.0026} & \textbf{0.9208$\pm$0.0003} & \textbf{8.2132$\pm$0.0030} & \textbf{38.5204$\pm$0.0938} & \textbf{0.9535$\pm$0.0013} & \textbf{0.0752$\pm$0.0009} \\
\midrule
\multicolumn{10}{l}{\textit{Published reference results: DASMamba study}} \\
\addlinespace[2pt]
NAFNet$^*$ \citep{nafnet} & 31.9078 & 0.9376 & 29.5394 & 33.7848 & 0.9200 & 8.3742 & 37.2672 & 0.9453 & 0.0859 \\
CODEFormer$^*$ \citep{codeformer} & 31.8879 & 0.9371 & 29.5584 & 33.8066 & 0.9203 & 8.3532 & 37.2846 & 0.9448 & 0.0856 \\
XFormer$^*$ \citep{xformer} & 31.9122 & 0.9373 & 29.4700 & 33.7610 & 0.9200 & 8.3947 & 37.2542 & 0.9450 & 0.0859 \\
Histoformer$^*$ \citep{ref12pdf} & 31.9210 & 0.9378 & 29.4177 & 33.8115 & 0.9206 & 8.3480 & 37.3628 & 0.9472 & 0.0850 \\
VmambaIR$^*$ \citep{vmambair} & 31.9277 & 0.9380 & 29.3439 & 33.7627 & 0.9203 & 8.3943 & 37.3673 & 0.9468 & 0.0847 \\
DASMamba-S \citep{dasmamba} & 31.7919 & 0.9355 & 29.8132 & 33.7948 & 0.9202 & 8.3646 & 37.2895 & 0.9460 & 0.0857 \\
\midrule
DASMamba (reproduced) \citep{dasmamba} & 32.1426 & 0.9450 & 28.6841 & 33.8374 & 0.9270 & 8.3262 & 37.3532 & 0.9395 & 0.0849 \\
ContiLNN-DASMamba (ours) & \textbf{32.5030$\pm$0.0438} & 0.9400$\pm$0.0016 & \textbf{27.5955$\pm$0.1229} & \textbf{34.0308$\pm$0.0163} & 0.9219$\pm$0.0003 & \textbf{8.1445$\pm$0.0142} & \textbf{38.4796$\pm$0.1090} & \textbf{0.9519$\pm$0.0014} & \textbf{0.0754$\pm$0.0007} \\
\bottomrule
\end{tabular}%
}
\end{table*}
\section{Experiments}
\subsection{Datasets}
We evaluate anatomical restoration across three modalities with distinct degradations: CT denoising, MRI super-resolution and reduced-count PET restoration. The case-disjoint partitions and preprocessing follow the established restoration benchmarks \citep{restoreRWKV,dasmamba}. Low-quality (LQ) and high-quality (HQ) images denote the degraded input and paired reference, respectively.

The CT benchmark derives from the ten-patient training set released for the American Association of Physicists in Medicine (AAPM) Low Dose CT Grand Challenge, which provides paired quarter-dose and normal-dose images \citep{ldctChallenge}. The training, validation and test partitions contain eight, one and one patients, respectively, with 2,039, 128 and 211 slices. Quarter-dose images serve as inputs and normal-dose images as references.

For MRI, we use T2-weighted images from IXI \citep{ixiDataset}. The source allocation is 405/59/114 subjects for training/validation/testing, yielding 40,500/5,828/11,400 available slices; the validation reader supplies 58 readable subjects, as detailed in Supplementary Section~A. LQ images are reconstructed after retaining the central 6.25\% of two-dimensional k-space, corresponding to four-fold degradation along each in-plane axis. The slice-axis sampling is retained in this super-resolution task.

The PET cohort comprises 159 scans acquired with a PolarStar m660 system, originating from the dataset described by \citet{taskadaptiveRoutingMIR}. The training, validation and test partitions contain 120, ten and 29 scans, respectively. After exclusion of air-only slices, these partitions contain 8,350/684/2,044 paired images. Reduced-count inputs are generated from list-mode data with a reduction factor of 12. Modality-specific intensity normalization and evaluation windows are specified in Supplementary Section~A.

\subsection{Experimental setup}
We reproduce the 2D backbones using their released training procedures and initialize ContiLNN from the checkpoint with the highest validation score. A frozen copy of the corresponding backbone supplies predictions for 2D backbone distillation, while the ContiLNN backbone and Bi-CfC modules are trained jointly. As illustrated in Fig.~\ref{fig:architecture}, the architecture augments in-plane feature extraction with slice-axis propagation and uses the frozen backbone only to guide training. Contiguous slice windows are used for the principal restoration results in Table~\ref{tab:main_results} and the operator comparisons in Tables~\ref{tab:mixed_gap_robustness} and~\ref{tab:main_ablation}; mixed-gap training is investigated separately in Supplementary Section~H. The representative ContiLNN-RWKV optimization settings, window configurations and loss weights are collected in Supplementary Section~A, together with the common workflow used for DASMamba.

Restoration quality is measured using peak signal-to-noise ratio (PSNR), the structural similarity index measure (SSIM) \citep{ssim}, and root mean squared error (RMSE). Metrics are calculated in the modality's restored intensity scale, averaged within each patient or scan, and then averaged across cases with equal weight. ContiLNN estimates in Table~\ref{tab:main_results} are means and sample standard deviations (SDs) across five training seeds. Case-level statistical analyses are presented separately below. CT results describe the single held-out patient in its benchmark partition.

% Editor: Jialei.He
\begin{table*}[!t]
\centering
\caption{Bi-CfC and Bi-GRU under contiguous training on the 29-case PET test cohort. Both variants use the same insertion sites, objective and training budget. The slice-index gap $g$ specifies the separation between neighboring observations; irregular windows mix $g\in\{1,2,3\}$. All settings evaluate the same original target slices with different neighboring context. Bold indicates the better operator within each test condition.}
\label{tab:mixed_gap_robustness}
\small
\setlength{\tabcolsep}{14pt}
\renewcommand{\arraystretch}{1.08}
\begin{tabular}{llccc}
\toprule
Test context & Slice-axis operator & PSNR $\uparrow$ & SSIM $\uparrow$ & RMSE $\downarrow$ \\
\midrule
\multirow{2}{*}{Gap $g=1$} & Bi-GRU & 38.5176 & 0.9538 & 0.0753 \\
 & Bi-CfC & \textbf{38.5868} & \textbf{0.9546} & \textbf{0.0744} \\
\midrule
\multirow{2}{*}{Gap $g=2$} & Bi-GRU & 37.6217 & 0.9462 & 0.0826 \\
 & Bi-CfC & \textbf{37.7136} & \textbf{0.9476} & \textbf{0.0817} \\
\midrule
\multirow{2}{*}{Gap $g=3$} & Bi-GRU & 36.5295 & 0.9366 & 0.0940 \\
 & Bi-CfC & \textbf{36.6652} & \textbf{0.9393} & \textbf{0.0927} \\
\midrule
\multirow{2}{*}{Irregular gaps $\{1,2,3\}$} & Bi-GRU & 37.8544 & 0.9465 & 0.0802 \\
 & Bi-CfC & \textbf{37.9494} & \textbf{0.9479} & \textbf{0.0793} \\
\bottomrule
\end{tabular}
\end{table*}

\subsection{Quantitative results}
Ordered slice-axis modeling improves fidelity across the three evaluated restoration tasks (Table~\ref{tab:main_results}). Relative to Restore-RWKV, mean PSNR increases by 0.1907, 1.0176 and 1.2482 dB on CT, MRI and PET, respectively, accompanied by higher SSIM and lower RMSE. The agreement across these metrics connects the use of neighboring anatomy with improved intensity accuracy and structural fidelity. On the held-out CT patient, the improvement is consistent across training seeds, with a PSNR SD of 0.0026 dB.

The improvements persist across the observed seed variation. Every ContiLNN-RWKV training seed yields higher PSNR and SSIM and lower RMSE than the baseline in each modality. At one SD below the mean, PSNR remains 0.19, 0.99 and 1.15 dB above the baseline for CT, MRI and PET, respectively. This separation indicates that the benefit of neighboring anatomical context is maintained across the evaluated initializations.

Integration with DASMamba supports the use of slice-axis dynamics with a second in-plane representation. Mean PSNR increases by 0.1934, 0.3604 and 1.1264 dB on CT, MRI and PET, respectively, with corresponding RMSE reductions of 0.1817, 1.0886 and 0.0095 (Table~\ref{tab:main_results}). PET also improves in SSIM, while the CT and MRI SSIM estimates are lower than those of their DASMamba baselines by 0.0051 and 0.0050, respectively. The consistent PSNR and RMSE improvements across both backbones support architectural compatibility, with the accompanying SSIM differences retained in the paired results.

\subsection{Sampling robustness}
Bi-CfC is evaluated against Bi-GRU to examine restoration when neighboring slices are sparsely or irregularly available (Table~\ref{tab:mixed_gap_robustness}). Both variants are trained on contiguous windows, retaining the same backbone initialization, insertion sites, full objective and training budget. At test time, sampling changes the neighboring context while preserving the original images, slice indices and evaluated target slices.

Uniform gaps $g=1,2,3$ select consecutive, every second and every third slices, respectively. For full seven-slice irregular windows, the six transitions comprise two occurrences of each gap. Sparse evaluation uses interleaved slice sequences, whereas irregular evaluation combines overlapping windows; both protocols retain the same eligible targets across the 29 test cases.

Under contiguous training, Bi-CfC achieves higher PSNR and SSIM and lower RMSE than Bi-GRU in all four test contexts (Table~\ref{tab:mixed_gap_robustness}; Supplementary Fig.~S6A). Its PSNR advantage increases from 0.0693 dB at $g=1$ to 0.1357 dB at $g=3$. Relative to each operator's $g=1$ result, PSNR decreases by 1.9216 dB for Bi-CfC and 1.9881 dB for Bi-GRU at $g=3$, indicating modestly smaller degradation as neighboring observations become more widely separated. Bi-CfC also has lower measured latency and peak allocated memory than Bi-GRU under the common profiling workload (Table~\ref{tab:params}; Supplementary Fig.~S6B).

An additional mixed-gap study examines whether exposure to varied sampling gaps improves restoration under altered neighboring context. Both operators benefit under sparse and irregular conditions, with no uniform ranking across metrics and contexts after mixed-gap training (Supplementary Section~H).

\begin{figure*}[!tp]
\centering
\includegraphics[width=0.99\textwidth]{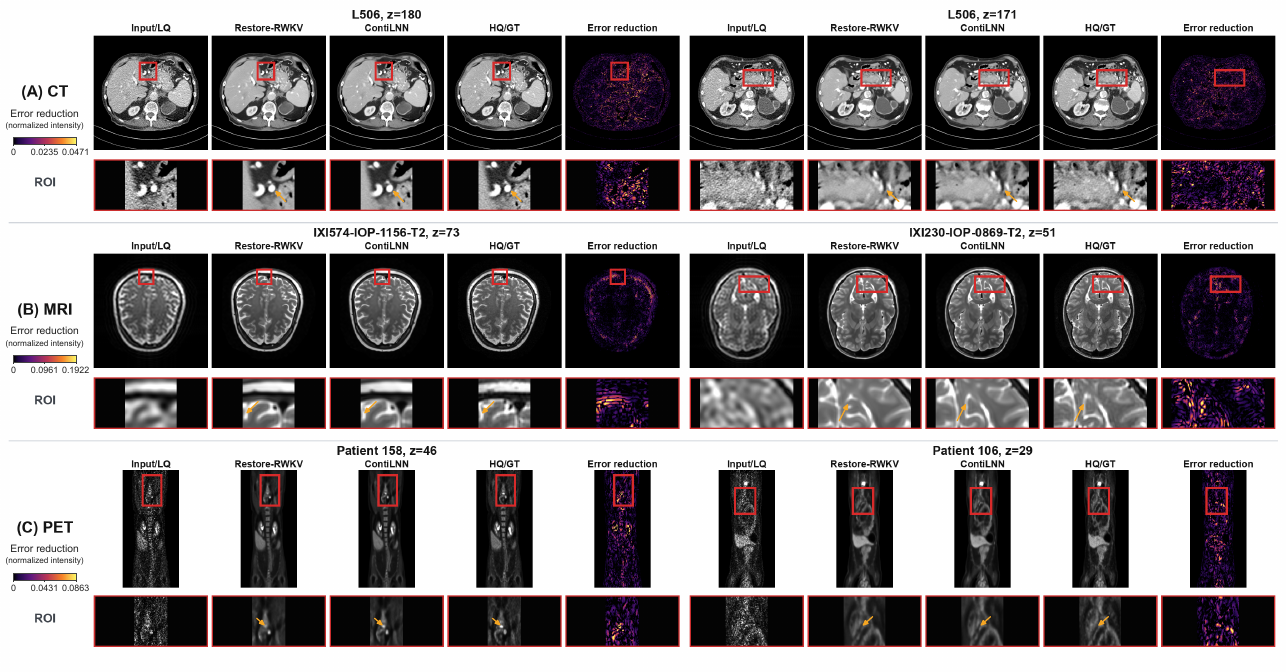}
\caption{Restoration examples for (A) CT, (B) MRI and (C) PET, with two examples per modality. Columns contain LQ inputs, Restore-RWKV and ContiLNN outputs, HQ references (ground truth, GT), and error-reduction maps. Boxed regions are enlarged below; orange arrows identify corresponding local structures and intensity transitions. Brighter colors indicate larger positive reductions in absolute error relative to Restore-RWKV, measured in normalized display intensity. Error increases are clipped to zero. Each modality shares one color scale; display masking and saturation settings are given in Supplementary Section~A.}
\label{fig:qualitative_comparison}
\end{figure*}

% Editor: Jialei.He
\begin{table*}[!t]
\centering
\caption{PET restoration with 5\% and 10\% of the training patients. Patient subsets are nested, with the validation and test cohorts unchanged. Budget-matched training uses the same number of slice presentations as Restore-RWKV; full refinement adds optimization on the same patient subset. Bold highlights the ContiLNN results.}
\label{tab:data_hunger}
\small
\setlength{\tabcolsep}{5pt}
\renewcommand{\arraystretch}{1.12}
\begin{tabular}{llccc}
\toprule
Patient budget & Training strategy & PSNR $\uparrow$ & SSIM $\uparrow$ & RMSE $\downarrow$ \\
\midrule
5\% & Restore-RWKV & 36.2923 & 0.9372 & 0.0973 \\
5\% & Budget-matched ContiLNN & \textbf{36.8837} & \textbf{0.9390} & \textbf{0.0911} \\
5\% & Full-refinement ContiLNN & \textbf{37.2385} & \textbf{0.9343} & \textbf{0.0875} \\
\midrule
10\% & Restore-RWKV & 36.5217 & 0.9389 & 0.0947 \\
10\% & Budget-matched ContiLNN & \textbf{37.1106} & \textbf{0.9408} & \textbf{0.0886} \\
10\% & Full-refinement ContiLNN & \textbf{37.2121} & \textbf{0.9430} & \textbf{0.0875} \\
\bottomrule
\end{tabular}
\end{table*}

\subsection{Data efficiency}
Limited paired cohorts provide a second test of the information contributed by neighboring anatomy. We train PET models using nested subsets of six and 12 patients, corresponding to 5\% and 10\% of the training cohort, while retaining the full validation and test partitions (Table~\ref{tab:data_hunger}). Budget-matched training assigns the same total number of target-slice presentations to Restore-RWKV and ContiLNN. Full refinement additionally trains the sequence model from the corresponding 2D checkpoint; both strategies use the same patients. Exposure budgets and schedules are detailed in Supplementary Section~A.

Under the matched target-slice exposure budget, ContiLNN improves PSNR by 0.5914 and 0.5889 dB at the 5\% and 10\% patient fractions, respectively, with higher SSIM and lower RMSE. Full refinement uses additional optimization on the same patient subsets and yields further PSNR and RMSE improvements. Its SSIM decreases at 5\% and increases at 10\% relative to matched-budget ContiLNN. The matched-budget results support improved use of limited paired data, while the separate refinement results describe the effect of additional optimization.

\begin{figure*}[!tp]
\centering
\includegraphics[width=0.94\textwidth]{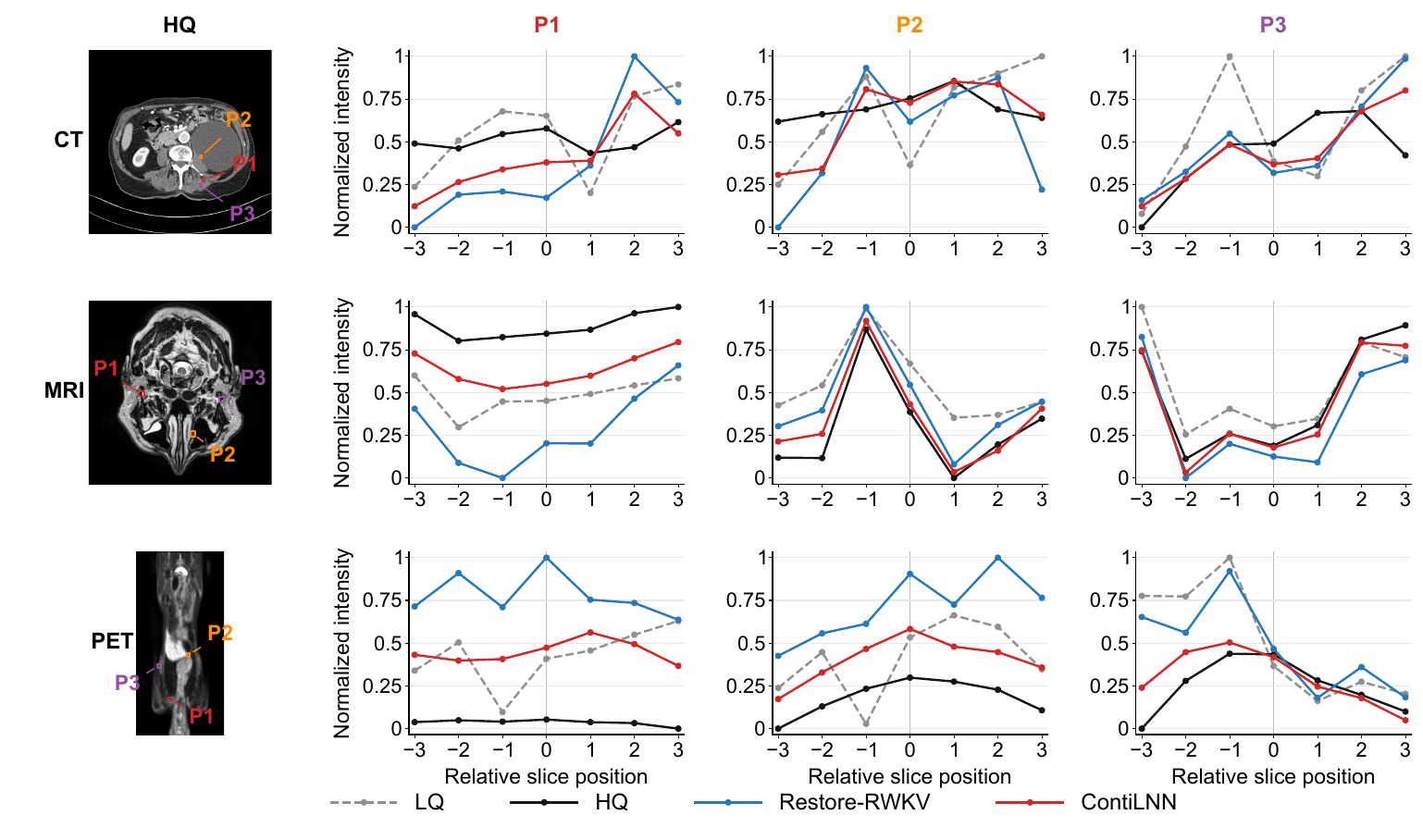}
\caption{Intensity profiles across seven ordered CT, MRI and PET slices. P1--P3 mark tissue interfaces or uptake transitions. Each profile averages a $7\times7$ region centered at the marked location and is normalized using the joint range of the LQ, HQ, Restore-RWKV and ContiLNN profiles at that location. The horizontal axis gives relative slice position; colors and line styles identify the four image sequences.}
\label{fig:z_profile_consistency}
\end{figure*}
\subsection{Qualitative results}
ContiLNN preserves local structure across the three modalities. The CT examples retain clearer soft-tissue interfaces and small high-contrast structures (Fig.~\ref{fig:qualitative_comparison}A). MRI contours follow the HQ reference more closely (Fig.~\ref{fig:qualitative_comparison}B), while PET uptake patterns show clearer local contrast (Fig.~\ref{fig:qualitative_comparison}C). The matched regions of interest and error maps locate these improvements.

Orthogonal MRI views extend this assessment beyond individual in-plane slices (Supplementary Section~G, Figs.~S4 and S5). The enlarged regions highlight narrow tissue interfaces and branching boundaries in both the H--W and D-axis views; ContiLNN follows their HQ appearance more closely than Restore-RWKV in these selected regions. The D--W and D--H reformats retain all 100 processed slices, allowing these local structures to be examined across the available slice sequence.

Boundary regions provide a focused assessment of cross-slice continuity: as anatomy varies along the slice axis, the relative contributions of neighboring tissues within a fixed local region change, producing intensity transitions that a restoration model should preserve. We examine three locations (P1--P3) in each modality (Fig.~\ref{fig:z_profile_consistency}). In CT, P1 and P3 lie near posterior paraspinal soft-tissue boundaries, while P2 lies beside the vertebral body at a soft-tissue interface. In MRI, P1 and P3 sample soft-tissue interfaces near the skull base, and P2 samples a tissue--air interface in the nasal region. In PET, P1--P3 sample pelvic, abdominal and lateral body regions with distinct uptake contrasts. The profiles thus probe both relatively stable local intensities and more pronounced changes as structures pass through the sampled region.

Agreement with the reference involves preserving the level and shape of these intensity changes. At CT P2, ContiLNN reduces the intensity deviation across all seven positions, including the terminal drop in the baseline profile. At MRI P2, it follows the reference peak at relative position $-1$ and trough at $+1$, retaining the sharp transition. At PET P3, it more closely follows the rise and subsequent decline in reference uptake. For these three displayed profiles, the mean absolute deviation from the reference over seven positions decreases from 0.290 to 0.135, from 0.150 to 0.066 and from 0.257 to 0.085, respectively, on the normalized intensity scale defined in the caption. Together, these examples link the visible boundary regions to reduced local intensity error and better preservation of reference variation along the slice axis.

\begin{figure*}[!t]
\centering
\includegraphics[width=0.96\textwidth]{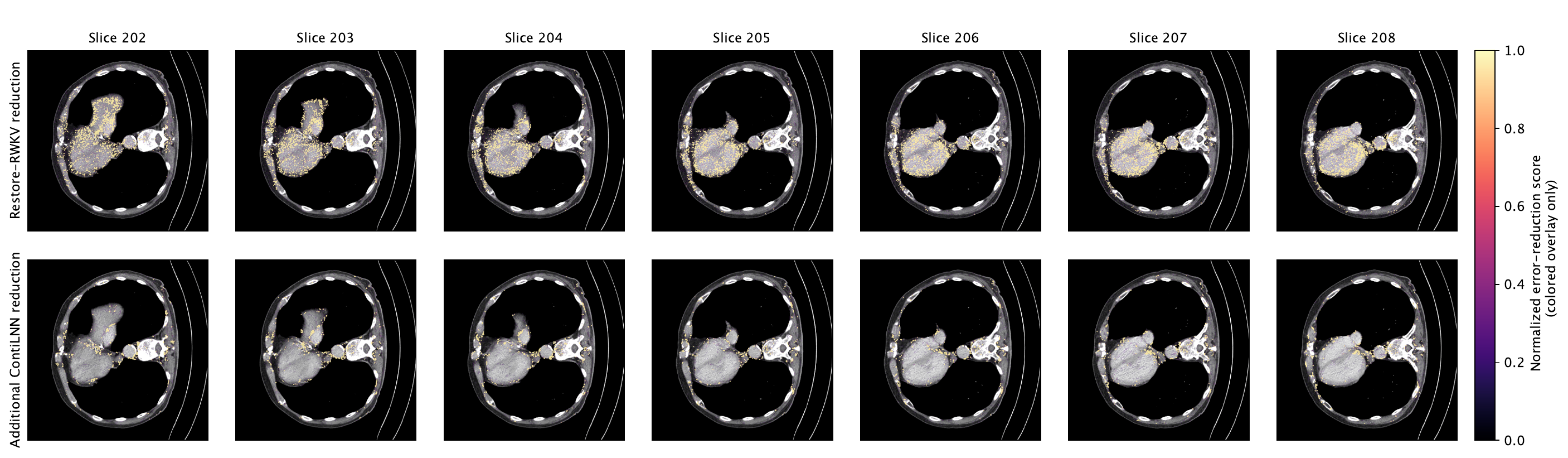}
\caption{Spatial error reduction across seven adjacent CT slices: Restore-RWKV relative to LQ (top), and ContiLNN relative to Restore-RWKV (bottom). Colored overlays indicate positive reductions in absolute error against HQ; each row shares one color scale.}
\label{fig:spatial_gain_sequence}
\end{figure*}
Additional refinement is visible along the lung-facing margins of central thoracic soft tissue and near posterior structures beside the vertebral body (Fig.~\ref{fig:spatial_gain_sequence}, lower row). Across slices 202--208, the highlighted regions remain associated with these interfaces as their contours change. Relative to the LQ input, Restore-RWKV reduces error across broader soft-tissue regions (upper row), with further reductions near anatomical boundaries after ContiLNN refinement (lower row). The recurring reductions at corresponding interfaces complement the intensity variations at the selected locations in Fig.~\ref{fig:z_profile_consistency}, providing spatial and local profile evidence of restoration along the slice sequence. This pattern is consistent with neighboring anatomical information contributing to local boundary refinement.

\begin{figure*}[!t]
\centering
\includegraphics[width=0.96\textwidth]{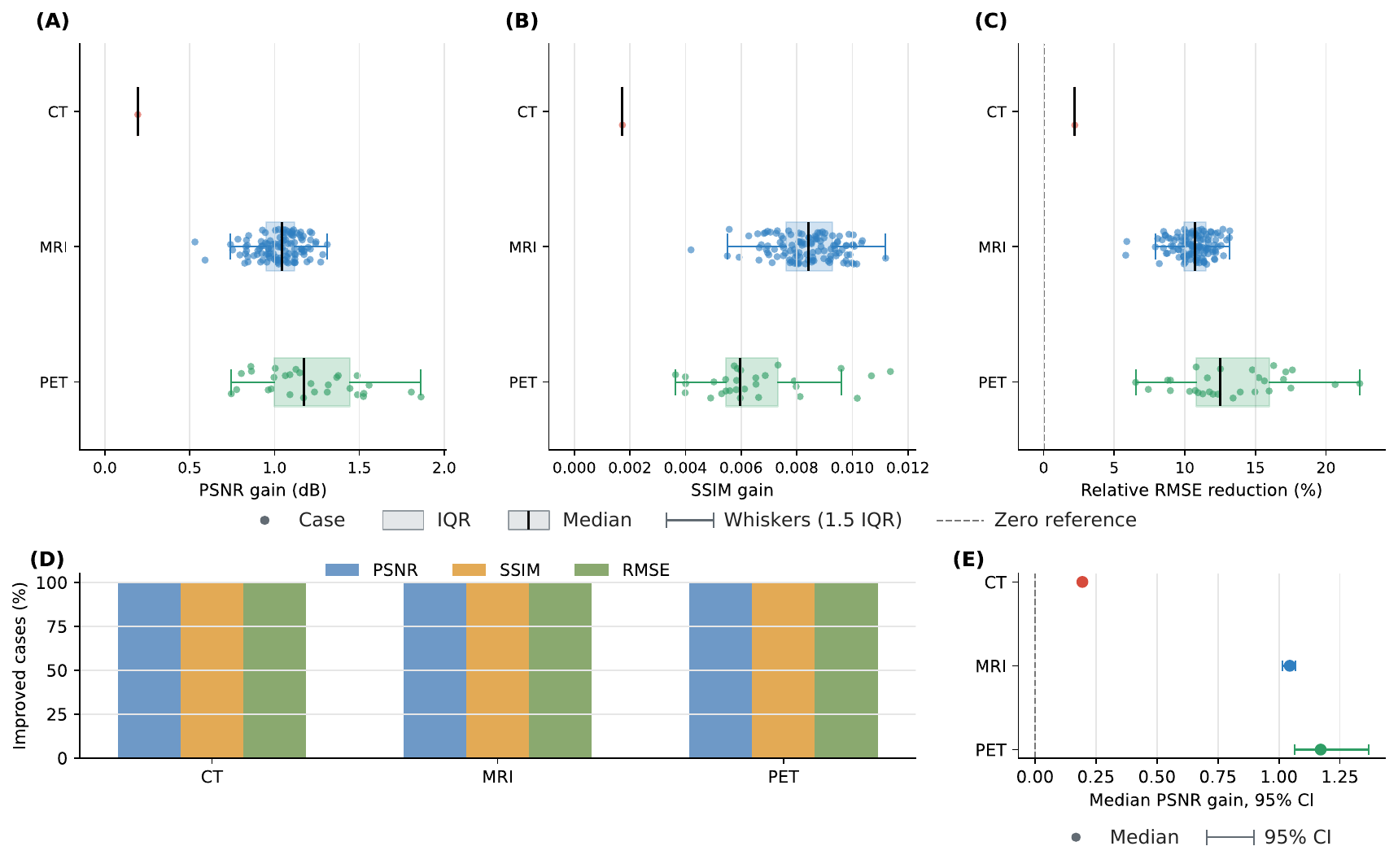}
\caption{Case-level improvements over Restore-RWKV. (A)--(C) PSNR gain, SSIM gain and relative RMSE reduction; positive values favor ContiLNN. Points show all cases. Boxes span the interquartile range (IQR), with median lines and whiskers ending at the most extreme observations within 1.5 IQR of the box. Dashed lines mark zero. (D) Percentage of improved cases. (E) Median PSNR gains with 95\% percentile confidence intervals from 10,000 case-bootstrap resamples for MRI and PET. CT has one test case and is descriptive, without a confidence interval.}
\label{fig:paired_statistics}
\end{figure*}
% Editor: Jialei.He
\begin{table*}[!t]
\centering
\caption{Computational cost and PET restoration gains. Profiling uses $128\times128$ images, 32-bit precision and one NVIDIA A800 GPU: seven independent slices for Restore-RWKV ($B=7$), or one seven-slice window for each ContiLNN-RWKV variant. MLP and Bi-GRU denote multilayer perceptron and bidirectional gated recurrent unit. M denotes $10^6$ parameters; GMAC denotes $10^9$ multiply--accumulate operations. Latency is median forward time per slice; memory is peak allocated GPU memory. Each $\Delta$ is the model's result minus the reproduced Restore-RWKV result in Table~\ref{tab:main_ablation} (37.2722 dB PSNR, 0.9474 SSIM, 0.0860 RMSE). Positive PSNR/SSIM and negative RMSE differences indicate improvement. Profiling details are in Supplementary Section~A.}
\label{tab:params}
\small
\setlength{\tabcolsep}{3pt}
\renewcommand{\arraystretch}{1.12}
\begin{tabular}{@{}lcccccccc@{}}
\toprule
& \multicolumn{5}{c}{Computational cost} & \multicolumn{3}{c}{PET metric differences} \\
\cmidrule(lr){2-6}\cmidrule(l){7-9}
Model & \shortstack{Params.\\(M)} & \shortstack{GMAC/\\slice} & \shortstack{GMAC/\\window} & \shortstack{Latency/\\slice (ms)} & \shortstack{Memory\\(MB)} & \shortstack{$\Delta$PSNR$\uparrow$\\(dB)} & $\Delta$SSIM$\uparrow$ & $\Delta$RMSE$\downarrow$ \\
\midrule
Restore-RWKV & 28.37 & 37.46 & 262.21 & 16.72 & 987.00 & 0.0000 & 0.0000 & 0.0000 \\
ContiLNN-RWKV, MLP & 34.92 & 60.32 & 422.26 & 21.23 & 1835.22 & $-0.3039$ & $-0.0039$ & $+0.0031$ \\
ContiLNN-RWKV, Bi-GRU & 34.91 & 60.50 & 423.51 & 54.42 & 2315.80 & $+1.2454$ & $+0.0064$ & $-0.0107$ \\
ContiLNN-RWKV, Bi-CfC & 34.93 & 60.34 & 422.39 & 25.69 & 995.03 & $+1.3146$ & $+0.0072$ & $-0.0116$ \\
\bottomrule
\end{tabular}
\end{table*}

\subsection{Case-level analysis}
Restoration gains extend across the held-out MRI and PET cohorts. We analyze paired case-level differences from complete saved output volumes, defining positive differences as higher PSNR or SSIM and lower RMSE. Significance is evaluated with paired two-sided Wilcoxon signed-rank tests \citep{wilcoxon1945} and Holm adjustment across the six modality--metric tests \citep{holm1979}. Median differences are accompanied by 95\% case-bootstrap confidence intervals \citep{efron1979}; the evaluation settings are given in Supplementary Section~A.

All 114 MRI subjects and all 29 PET scans improve in PSNR, SSIM and RMSE (Fig.~\ref{fig:paired_statistics}A--D). Median PSNR gains are 1.0443 dB for MRI and 1.1711 dB for PET, with 95\% confidence intervals of [1.0154, 1.0681] and [1.0636, 1.3666] dB, respectively (Fig.~\ref{fig:paired_statistics}E). Median SSIM increases are 0.0084 and 0.0060, and median absolute RMSE reductions are 2.0425 and 0.0104. All six adjusted $p$ values are below $1.2\times10^{-8}$. The distributions therefore associate the aggregate improvement with consistent case-level benefits throughout these two test cohorts.

% Editor: Jialei.He
\begin{table*}[!t]
\centering
\caption{PET ablation on the 29-case test cohort. ContiLNN variants share the pretrained backbone initialization, insertion locations, training budget and evaluation procedure. The MLP processes slices independently; Bi-GRU and Bi-CfC use the same bidirectional slice context. $\mathcal{L}_{\mathrm{full}}$ denotes the full training objective. Bold indicates the highest PSNR and SSIM and the lowest RMSE.}
\label{tab:main_ablation}
\small
\setlength{\tabcolsep}{4pt}
\renewcommand{\arraystretch}{1.12}
\begin{tabular}{llllccc}
\toprule
ID & Setting & Slice-axis operator & Objective & PSNR $\uparrow$ & SSIM $\uparrow$ & RMSE $\downarrow$ \\
\midrule
A & Restore-RWKV & None & $\mathcal{L}_{\mathrm{rec}}$ & 37.2722 & 0.9474 & 0.0860 \\
B & ContiLNN, reconstruction only & Bi-CfC & $\mathcal{L}_{\mathrm{rec}}$ & 38.4166 & 0.9533 & 0.0760 \\
C & ContiLNN, pointwise MLP & Slice-independent MLP & $\mathcal{L}_{\mathrm{full}}$ & 36.9683 & 0.9435 & 0.0891 \\
D & ContiLNN, recurrent alternative & Bi-GRU & $\mathcal{L}_{\mathrm{full}}$ & 38.5176 & 0.9538 & 0.0753 \\
E & ContiLNN, full objective & Bi-CfC & $\mathcal{L}_{\mathrm{full}}$ & \textbf{38.5868} & \textbf{0.9546} & \textbf{0.0744} \\
\bottomrule
\end{tabular}
\end{table*}

\subsection{Ablation studies}
\label{sec:ablations}
\paragraph{Computational cost}
Selective insertion adds 6.56 million parameters and 22.88 billion multiply--accumulate operations (GMACs) per slice to Restore-RWKV (Table~\ref{tab:params}). For a matched seven-image workload, the median forward time per slice increases by 8.97 ms, from 16.72 to 25.69 ms, while peak allocated inference memory changes from 987.00 to 995.03 MB. Per-slice time is amortized over one seven-slice forward pass; data loading and overlapping-window fusion are outside this measurement. Profiling uses the same device, precision and image dimensions, specified in Supplementary Section~A.

A slice-independent multilayer perceptron (MLP) inserted at the same pathways has nearly identical model size and arithmetic cost to Bi-CfC. Its latency is lower, at 21.23 ms per slice, but its peak allocated memory is higher, at 1,835.22 MB. This resource profile enables an ablation of ordered propagation against a pointwise operator with closely matched capacity and computation.

A bidirectional gated recurrent unit (Bi-GRU) alternative is evaluated under the same profiling conditions. Bi-GRU and Bi-CfC closely match in parameter count (34.91 versus 34.93 million) and arithmetic cost (423.51 versus 422.39 GMACs per window). Bi-CfC reduces median window latency from 380.92 to 179.85 ms (52.8\%) and peak allocated memory from 2,315.80 to 995.03 MB (57.0\%), showing that matched capacity and arithmetic cost do not imply matched execution cost.

\paragraph{Architecture and objective}
The PET ablations examine pointwise capacity, ordered-operator choice and training regularization (Table~\ref{tab:main_ablation}). Reconstruction-only Bi-CfC improves PSNR by 1.1444 dB over the 2D backbone. With the full objective, it exceeds the slice-independent MLP by 1.6185 dB in PSNR and 0.0111 in SSIM, with RMSE reduced by 0.0147, supporting propagation beyond pointwise capacity. Bi-GRU retains the same bidirectional context, full objective, initialization, insertion sites and training budget. Relative to this recurrent alternative, Bi-CfC improves PSNR by 0.0693 dB and SSIM by 0.0008, reduces RMSE by 0.0009, and has lower measured latency and memory use (Table~\ref{tab:params}). Together with the sampling comparison in Table~\ref{tab:mixed_gap_robustness} and the resource measurements in Table~\ref{tab:params}, these results support Bi-CfC as the preferred sequence operator for ContiLNN under the evaluated conditions.

Consistency and distillation provide a further improvement to Bi-CfC: the full objective adds 0.1702 dB in PSNR and 0.0013 in SSIM, while reducing RMSE by 0.0016. The broader architecture study supports selective insertion at intermediate and bottleneck features, with gated updates along the remaining pathways (Supplementary Section~B). Component-wise loss analysis is provided in Section~C, and sequence length is examined in Section~D. The completed 2D training trajectory in Supplementary Section~F documents the initialization used for refinement.

\begin{figure*}[!t]
\centering
\includegraphics[width=0.99\textwidth]{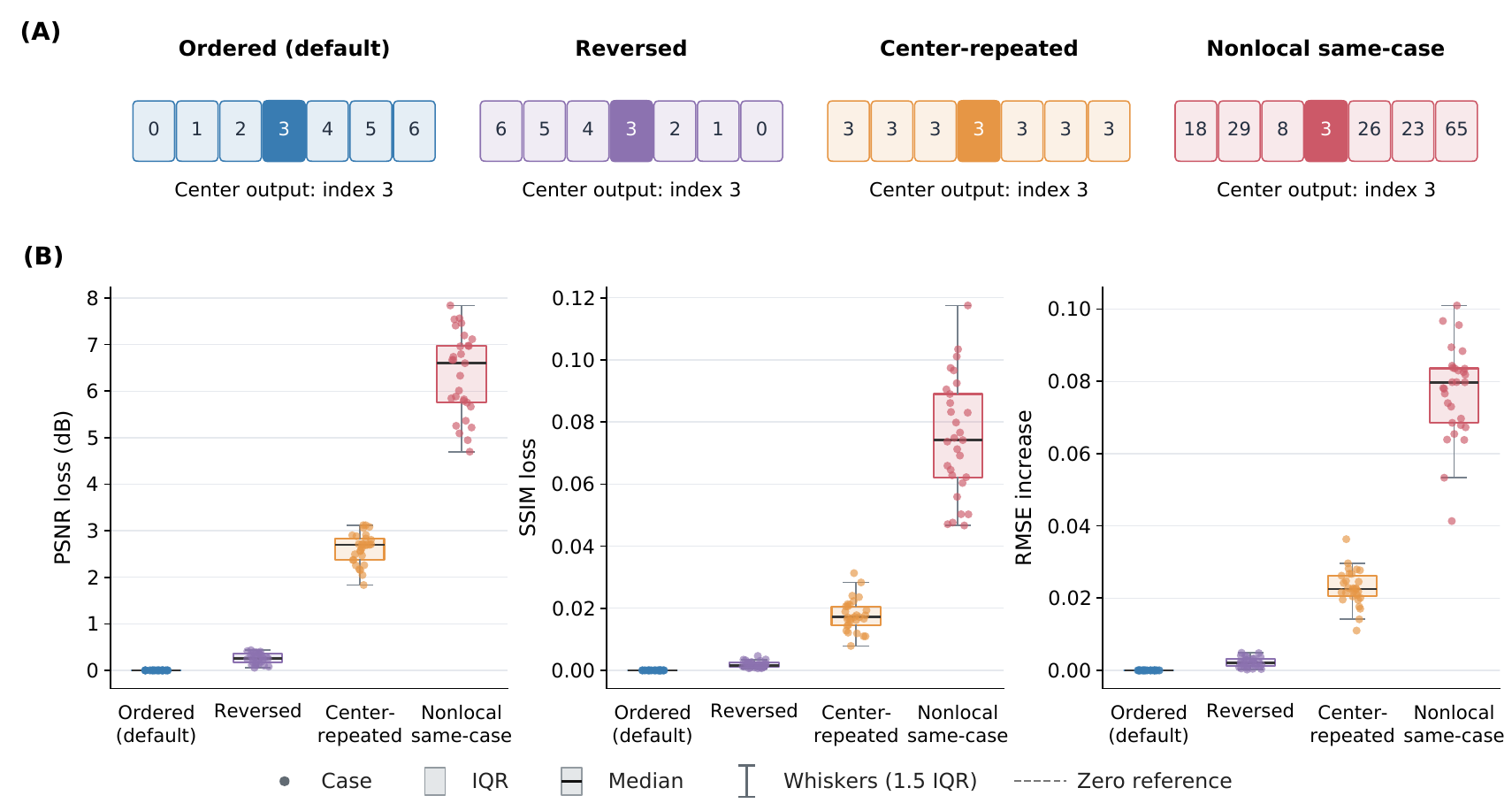}
\caption{PET context perturbations. (A) Ordered, reversed, center-repeated and nonlocal same-case inputs, illustrated using Patient 10 at center index 3; every condition retains the evaluated center image. (B) PSNR loss, SSIM loss and RMSE increase for 29 test scans using the same trained checkpoint. Points show all cases; box-plot conventions follow Fig.~\ref{fig:paired_statistics}A--C. Differences are relative to ordered context, whose zero value is marked by dashed lines.}
\label{fig:context_perturbation}
\end{figure*}
% Editor: Jialei.He
\begin{table*}[!t]
\centering
\caption{PET restoration under the context perturbations illustrated in Fig.~\ref{fig:context_perturbation}. The same trained ContiLNN-RWKV checkpoint is held fixed across all four conditions. Ten center slices are evaluated in each of the 29 test cases. Metrics are averaged over centers within each case and then across cases; $\Delta$PSNR is relative to ordered context. Bold indicates the best PSNR, SSIM and RMSE across these conditions.}
\label{tab:context_ablation}
\small
\setlength{\tabcolsep}{5pt}
\renewcommand{\arraystretch}{1.12}
\begin{tabular}{lrrrrrr}
\toprule
Context & Cases & Centers & PSNR $\uparrow$ & $\Delta$PSNR & SSIM $\uparrow$ & RMSE $\downarrow$ \\
\midrule
Ordered (default) & 29 & 290 & \textbf{38.719844} & 0.000000 & \textbf{0.954441} & \textbf{0.073956} \\
Reversed & 29 & 290 & 38.463333 & -0.256511 & 0.952564 & 0.076172 \\
Center-repeated & 29 & 290 & 36.104738 & -2.615106 & 0.936650 & 0.096815 \\
Nonlocal same-case & 29 & 290 & 32.368766 & -6.351078 & 0.879486 & 0.150989 \\
\bottomrule
\end{tabular}
\end{table*}
We assess local anatomy by varying context around a fixed PET center image, using the same trained ContiLNN-RWKV checkpoint (Fig.~\ref{fig:context_perturbation}A). Ordered local context is the default condition. Reversal retains the local images in the opposite order; center repetition removes distinct neighboring observations; nonlocal same-case sampling retains patient-specific appearance while replacing local anatomy. Each condition is evaluated at the same ten center positions per scan across all 29 test scans.

As shown in Fig.~\ref{fig:context_perturbation}B and the results in Table~\ref{tab:context_ablation}, reversing the local sequence reduces PSNR by 0.2565 dB. Preceding and following content remains available to the bidirectional operator, while its positional arrangement changes. Center repetition and nonlocal same-case sampling reduce PSNR by 2.6151 and 6.3511 dB, respectively. These changes associate restoration quality with both the locality and ordering of the supplied observations, with the larger deterioration following the removal of distinct local neighbors. Supplementary Section~E describes the paired analysis, and Supplementary Table S6 supplies the per-case absolute metrics and changes relative to ordered context.

% Editor: Jialei.He
\section{Discussion}
\subsection{Anatomical context and restoration}
Earlier restoration studies established complementary uses of adjacent slices through volumetric feature extraction and neighboring-slice supervision \citep{cpce3d,noise2context}. ContiLNN builds on this anatomical information through ordered feature propagation within a 2D restoration hierarchy. The PET ablations distinguish additional pointwise capacity from cross-slice propagation and compare two operators with the same bidirectional context. Context perturbations further show larger losses when distinct local neighbors are removed or replaced than when their order is reversed (Table~\ref{tab:context_ablation}). This pattern suggests that local anatomical content supplies an important source of information, with its ordering making an additional contribution to restoration fidelity.

The two regularizers address different aspects of using this information. Reference-based differences encourage predictions to follow the intensity changes of the anatomical sequence, including transitions at tissue interfaces. Knowledge distillation transfers information through reference predictions \citep{hintonKD}; here, 2D backbone distillation uses a frozen, pretrained backbone to provide an in-plane reference as ContiLNN incorporates neighboring observations. Their complementary role is supported by the loss-component analysis in Supplementary Section~C. The resulting design combines access to neighboring anatomy with supervision of how the restored sequence should vary.

Across CT, MRI and PET, this combination improves fidelity under different degradation processes. Local image examples, orthogonal MRI views and intensity profiles relate the gains to anatomical interfaces and uptake transitions, while the MRI and PET case-level analyses show how those gains extend across the evaluated cohorts. Together, these results support local anatomical continuity as a useful source of complementary information for restoration.

\subsection{Sampling and efficiency}
Restoration from incomplete image series requires using neighboring observations whose separation may vary along the slice axis. Retaining the original indices distinguishes consecutive slices from observations separated by omitted positions. CfC incorporates the resulting normalized index gaps into its closed-form state updates without iterative numerical integration \citep{cfc}. Under the principal contiguous-training protocol, Bi-CfC maintains higher restoration fidelity than Bi-GRU across the tested sparse and irregular contexts. The mixed-gap study shows that broader sampling exposure benefits both operators, with closely matched performance under that training strategy. Together with lower measured latency and peak allocated memory than Bi-GRU, these findings support Bi-CfC as the preferred slice-axis operator for ContiLNN in the evaluated setting.

The limited-patient PET experiments examine the value of the same context under constrained training data. Improvements at matched target-slice exposure indicate more effective use of the available observations. Additional refinement addresses a complementary setting by allocating further optimization to models trained on the same patients. The two settings connect sequence modeling with data efficiency while distinguishing the contribution obtained at a matched exposure budget from that obtained through continued training.

Integration with Restore-RWKV and DASMamba \citep{restoreRWKV,dasmamba} separates in-plane representation learning from cross-slice information exchange. PSNR and RMSE gains with both backbones support this division of roles across the two tested architectures. Operator choice also affects the resources required for this exchange. Under the matched seven-slice, 32-bit NVIDIA A800 profiling conditions, Bi-CfC reduces median window latency by 52.8\% and peak allocated memory by 57.0\% relative to Bi-GRU, despite closely matched parameter counts and arithmetic costs (Table~\ref{tab:params}). The limited-patient and operator studies therefore identify complementary benefits in the use of training data and inference resources.

\subsection{Limitations}
The current evidence is based on fixed restoration benchmarks and local slice windows. Physical slice-spacing metadata are unavailable in the preprocessed data used here, so the experiments assess index-defined sampling rather than adaptation to calibrated anatomical distances. Window length and index gaps jointly determine the anatomical extent of the available context.

The CT test partition contains one patient; the small variation across training seeds characterizes training repeatability on that patient rather than between-patient robustness. Generalization across acquisition protocols has not yet been evaluated. Adding slice-axis propagation also increases the parameter count and arithmetic cost relative to the original 2D backbone.

\subsection{Future directions}
Physical slice coordinates derived from Digital Imaging and Communications in Medicine (DICOM) metadata could connect the updates more directly to anatomical distance and acquisition geometry. Adaptive context lengths and selective propagation across feature levels could further match contextual exchange to sampling density and local anatomical variation. Comparisons with slice-axis convolution and attention under matched budgets would help characterize these choices. Evaluation across centers and acquisition protocols, together with downstream tasks and reader studies, would extend the assessment of restoration quality to broader imaging workflows. By improving the quality of available slices while preserving their original positions, ContiLNN may help recover useful anatomical information from incomplete image series in retrospective clinical workflows.

\section{Conclusion}
ContiLNN augments two-dimensional medical image restoration with bidirectional liquid dynamics over ordered slice positions. Slice-index intervals condition state propagation on the supplied sampling, while joint restoration preserves an output for every observed slice. Experiments across CT, MRI and PET demonstrate improved fidelity. Experiments with limited paired training cohorts (5\% and 10\% of PET training patients) and a second restoration backbone (DASMamba) support data efficiency and architectural compatibility, while the PET ablations support ordered propagation and complementary regularization. Under contiguous training, Bi-CfC achieves higher fidelity than Bi-GRU across the tested sampling conditions, alongside lower measured inference latency and peak allocated memory. Additional mixed-gap experiments show that broader sampling exposure can improve both operators. These findings support Bi-CfC as the preferred operator in the evaluated ContiLNN framework and liquid dynamics as a practical approach to incorporating anatomical continuity into 2D restoration networks.

% Editor: Jialei.He
\section*{Data availability}
The CT data derive from the ten-patient training set of the AAPM Low Dose CT Grand Challenge \citep{ldctChallenge}, and the MRI data from the IXI project \citep{ixiDataset}. The PolarStar m660 PET cohort is described in the original AMIR study \citep{taskadaptiveRoutingMIR}. The preprocessed benchmark datasets used in this study were obtained from the Restore-RWKV repository (\url{https://github.com/Yaziwel/Restore-RWKV}); their source cohorts and preprocessing are described in Section~4.1. Access and reuse are subject to the terms of the respective source data providers. No new clinical imaging data were collected.

\section*{Code availability}
The source code, experiment configurations, and evaluation scripts are publicly available at \url{https://github.com/Noah-hjl/ContiLNN}.

\section*{Declaration of generative AI and AI-assisted technologies in the manuscript preparation process}
During the preparation of this manuscript, the authors used ChatGPT and Codex (OpenAI) for language editing, manuscript organization, LaTeX preparation, literature checking, and assistance with plotting and layout code. The authors reviewed and verified the resulting text, references, and relevant code outputs, edited the material as needed, and take full responsibility for the content of the published article. Medical images presented as evidence originate from research data and model outputs, not generative AI.

\section*{Ethics statement}
This work involved secondary analysis of pre-existing data released for research use and included no new participant recruitment or imaging acquisition. The source cohorts and acquisition protocols are described in the cited dataset publications.

\section*{Declaration of competing interest}
Given his role as Associate Editor of Medical Image Analysis, S. Kevin Zhou had no involvement in the peer review of this article and had no access to information regarding its peer review. Full responsibility for the editorial process for this article was delegated to another journal editor.

\section*{Funding}
This research received no specific grant from any funding agency in the public, commercial, or not-for-profit sectors.

\section*{Acknowledgements}
The authors acknowledge the High-Performance Computing Platform of Xi'an Jiaotong-Liverpool University for providing the computational resources used in this study.

\bibliographystyle{elsarticle-harv}
% Mark the source dataset for editorial identification without changing
% the official bibliography style or its author-year citation label.
\begingroup
\let\contilnnBibitem\bibitem
\renewcommand{\bibitem}[2][]{\contilnnBibitem[#1]{#2}\ifstrequal{#2}{ixiDataset}{[dataset]\space}{}}
\bibliography{references}
\endgroup
\end{document}

\maketitle

\appendix

\section{Implementation details}
\label{sec:supp_implementation}

\subsection{Slice windows and liquid modules}
We use ordered windows of at most $T=7$ slices and retain short terminal windows. Under the default contiguous strategy, training partitions each case into non-overlapping windows and shuffles their order between epochs. For a window containing $K\leq T$ observations with a constant slice-index gap $g$, the interval supplied to the liquid cell is $\delta=g/K$. Thus, consecutive slices use $g=1$ and $\delta=1/K$. The separate within-window position encoding is linearly spaced from 0 to 1 when $K>1$, and is 0 for a single-slice window. Normalization uses the actual $K$, including terminal windows. A fixed-gap single-slice window receives $\delta_0=g$. Interval inputs are not clipped to $[0,1]$. Position encoding specifies relative order, while each interval records the gap associated with that observation in the original forward order.

For irregular sampling, the first gap is repeated to supply an interval for the initial observation. Features and their associated intervals are reversed together for the backward branch. For example, indices $[0,1,3,6,7,9,12]$ give
\begin{equation}
\boldsymbol\delta=\frac{1}{7}[1,1,2,3,1,2,3],\qquad
\boldsymbol\delta^{\mathrm{rev}}=\frac{1}{7}[3,2,1,3,2,1,1],
\end{equation}
where $\boldsymbol\delta^{\mathrm{rev}}$ lists the interval inputs in backward traversal order. This convention preserves the sampling information paired with each feature; it does not recompute intervals from reverse-direction transitions between slice positions.

The Restore-RWKV configuration uses a base width of 48, encoder-stage block counts of $[4,6,6,8]$, and four refinement blocks. Six bidirectional closed-form continuous-time (Bi-CfC) modules follow the six encoder-L2 blocks, and six follow the decoder-L2 blocks. One module is placed on each of the L2 and L3 skip pathways, one at the bottleneck before decoding, and one after the refinement pathway. These placements give 16 Bi-CfC modules and 15 external gates. The bottleneck uses serial insertion; all other selected pathways use gated residual insertion. Gate logits initialize at 0, except for the final refinement gate, which initializes at $-4$.

Each directional CfC cell uses the default gated update, with one internal feed-forward layer of width $M_l=256$, no dropout and no sparsity mask. The scaled hyperbolic-tangent activation in the joint feature map is $\rho(a)=1.7159\tanh(0.666a)$, applied elementwise. The two candidate heads use ordinary hyperbolic-tangent activations; the interval coefficient and gate intercept are unrestricted affine outputs. The implemented gate is $\sigma(\delta_t\eta_t+\xi_t)$, and the state is $(1-\alpha_t)\odot c_t^{(1)}+\alpha_t\odot c_t^{(2)}$, following the default CfC parameterization described in the main manuscript. The internal width $M_l$ is distinct from the recurrent-state width $C_l$, which equals the feature-channel count at pathway $l$. Forward and backward cells have separate parameters. The position projection maps one scalar to $C_l$ channels, and the bidirectional fusion maps $2C_l$ channels to $C_l$; both affine projections include a learned bias. The source-code names \texttt{ff1}, \texttt{ff2}, \texttt{time\_a} and \texttt{time\_b} correspond to the two candidate heads, interval coefficient and gate intercept in the main formulation, respectively.

At inference, two window tilings begin at offsets 0 and 3 in the case's slice sequence. Each tiling retains its terminal window, and predictions at positions covered by both tilings are averaged. The original slice order is restored after fusion. All windows are formed within individual cases.

\subsection{Restore-RWKV training}
The 2D Restore-RWKV baseline is trained for 300,000 optimizer updates using Adam, a batch size of four slices, and an $L_1$ reconstruction loss. Cosine decay reduces the learning rate from $2\times10^{-4}$ to $10^{-6}$. The checkpoint with the highest validation peak signal-to-noise ratio (PSNR) initializes ContiLNN and supplies predictions from a frozen copy of the 2D backbone for distillation.

ContiLNN training uses $128\times128$ patches and three NVIDIA A800 GPUs, with one sequence per GPU. AdamW optimizes the complete model with zero weight decay. The initial learning rates are $2\times10^{-4}$ for the 2D backbone and Bi-CfC parameters and $5\times10^{-4}$ for the external gates, with cosine decay to $10^{-6}$. Training uses automatic mixed precision and gradient clipping at a norm of 1.0. CT and PET have a 60-epoch cap; MRI has a 100-epoch cap. Early stopping is enabled after 15 epochs, with patience 10 and a minimum validation-PSNR improvement of 0.0005 dB. The five-seed results in main Table~1 use seeds 42, 2026, 3407, 777, and 12345.

The full Restore-RWKV objective uses $\lambda_{\mathrm{cons}}=0.30$, $\beta_2=2.0$, and $\lambda_{\mathrm{KD}}=0.05$, with unit reconstruction weight. The first- and second-order consistency terms use finite differences between the selected observations. In the reconstruction-only PET ablation, both $\lambda_{\mathrm{cons}}$ and $\lambda_{\mathrm{KD}}$ are zero. The multilayer perceptron (MLP) and bidirectional gated recurrent unit (Bi-GRU) ablations use the full objective. Bi-GRU replaces Bi-CfC while retaining its insertion sites, initialization, training budget and evaluation workflow. The reference 2D backbone is frozen, while the ContiLNN backbone and added modules remain trainable.

The MRI source split allocates 405, 59, and 114 cases to training, validation, and testing. The implemented validation reader supplies 58 readable cases, and validation scores are calculated over those 58 cases. CT, MRI, and PET use their respective data readers and case-grouping rules.

\subsection{DASMamba integration}
ContiLNN-DASMamba follows the same backbone-initialization, ordered-window training, 2D backbone distillation and evaluation workflow as ContiLNN-RWKV. Its native spatial and Fourier-domain reconstruction terms are retained alongside the cross-slice consistency and distillation objectives. Detailed hyperparameters are reported for ContiLNN-RWKV as the representative implementation.

\subsection{Sampling robustness}
The PET sampling study compares Bi-CfC and Bi-GRU under contiguous and mixed-gap training. For each operator, the mixed-gap configuration retains the backbone initialization, insertion sites, objective, optimization settings and training budget of its contiguous-training counterpart. Each mixed-gap training window uses a constant gap selected from $\{1,2,3\}$. The three gaps contribute approximately equal numbers of windows, while the total windows and optimizer updates per epoch match contiguous training. A seven-observation window spans 7, 13, or 19 original slice positions for gaps 1, 2, or 3, respectively. Validation uses consecutive slices with the original two-tiling fusion. Sparse testing uses interleaved slice sequences, and irregular testing combines overlapping windows. All eligible original target slices are retained for evaluation across sampling conditions, with case-level metrics computed over the same test cohort. Mixed-gap training uses 16-bit mixed precision with a fixed loss scale of 128 and gradient clipping at 1.0. Complete results for both training strategies are reported in Section~\ref{sec:supp_mixed_gap}.

\subsection{Data efficiency}
The PET data-efficiency experiments use training seed 42 and nested subsets of 6 and 12 patients, corresponding to 5\% and 10\% of the 120 training patients. The validation and test cohorts remain fixed. A slice presentation is counted whenever one target slice contributes to training; this makes the exposure budget comparable between independent-slice and sequence training.

Restore-RWKV receives 1.2 million slice presentations from random initialization. Budget-matched ContiLNN uses 0.6 million presentations for the 2D pathway, then activates Bi-CfC for a further 0.6 million presentations. The transition uses the checkpoint at that fixed exposure, with a frozen copy supplying the distillation targets during sequence training. Full refinement starts from the ratio-matched Restore-RWKV checkpoint after its 1.2-million-presentation schedule and adds a sequence-training budget derived from the full-data refinement schedule.

Both sequence-training strategies use AdamW, a global batch of three sequences, and peak learning rates of $2\times10^{-4}$ for backbone and Bi-CfC parameters and $5\times10^{-4}$ for gates. The learning rates warm up over the first 5\% of the sequence-training exposure and then decay to $10^{-6}$. Consistency and distillation weights increase linearly over the first 10\% to 0.30 and 0.05, respectively; the second-order consistency weight is 2.0. The complete ContiLNN model remains trainable. Training uses a fixed mixed-precision loss scale of 128, gradient clipping at 1.0, and no early stopping. These schedules are indexed by slice exposure rather than epoch count.

\subsection{Intensity processing}
Training uses intensity ranges of $[-1024,3072]$ Hounsfield units (HU) for CT, $[0,4095]$ for MRI and $[0,20]$ for PET. Random $128\times128$ crops, rotations and flips are applied consistently across the slices in each training window. Evaluation follows inverse normalization and truncation to $[-160,240]$ HU for CT, $[0,4095]$ for MRI and $[0,20]$ for PET.

For the error-reduction maps in main Figure~3, $I$ denotes displayed grayscale intensity normalized to $[0,1]$. The plotted quantity is
\begin{equation}
E^{+}=\max\!\left(\lvert I_{\mathrm{RWKV}}-I_{\mathrm{HQ}}\rvert-
\lvert I_{\mathrm{ContiLNN}}-I_{\mathrm{HQ}}\rvert,0\right).
\label{eq:supp_display_error_reduction}
\end{equation}
The subscripts identify the two predictions and the HQ reference. Values are displayed where $0.035<I_{\mathrm{HQ}}<0.985$, with pixels outside this mask set to zero. Inside the mask, zero includes unchanged error and error increases clipped to zero. The two examples within each modality share one color scale, which saturates at the 99th percentile of their pooled positive ROI values. These display settings do not change the reconstruction metrics.

\subsection{Computational profiling}
Inference is profiled on an NVIDIA A800 GPU using 32-bit floating-point precision and $128\times128$ images. The baseline processes a batch of seven independent slices, while the Bi-CfC, MLP and Bi-GRU variants each process one window of seven slices. Timing is measured after 20 warm-up passes and summarized by the median of 100 repetitions, with GPU synchronization around each repetition. Per-slice latency and arithmetic cost are obtained by dividing the window totals by seven. These latencies measure an amortized model forward pass; they exclude data loading, window assembly and the overlapping-tiling fusion used for full-volume restoration. Peak allocated memory is recorded after warm-up; memory values labelled MB use $2^{20}$ bytes. One GMAC denotes $10^9$ multiply--accumulate operations. Peak reserved memory is 1,168 MB for Bi-CfC, 2,692 MB for the pointwise MLP and 3,000 MB for Bi-GRU; the main table reports allocated memory for all models. The Bi-GRU and Bi-CfC variants contain 34.50 and 34.52 million trainable parameters, respectively, of which 6.54 and 6.55 million belong to their ordered operators. Their median seven-slice window latencies are 380.92 and 179.85 ms, respectively.

\subsection{Evaluation and statistical settings}
The structural similarity index measure (SSIM) uses an $11\times11$ Gaussian window with standard deviation 1.5. The dynamic range used in the image-quality metrics is determined from the corresponding target slice. Slice-level metrics are averaged within each case before aggregation across cases. Multi-seed estimates report the mean and sample standard deviation across the five training seeds.

Paired case analyses use the models trained with seed 2026 for CT and MRI and seed 3407 for PET. Case-bootstrap intervals use 10,000 resamples and random seed 20260708. The statistical significance level is 0.05; the paired Wilcoxon tests, Holm adjustment, and confidence-interval interpretation are specified in the main manuscript. The CT test cohort contains one case, for which the reported measurements describe that case.

\section{Architecture ablation}
\label{sec:supp_architecture}
\begin{figure}[H]
\centering
\includegraphics[width=0.86\textwidth]{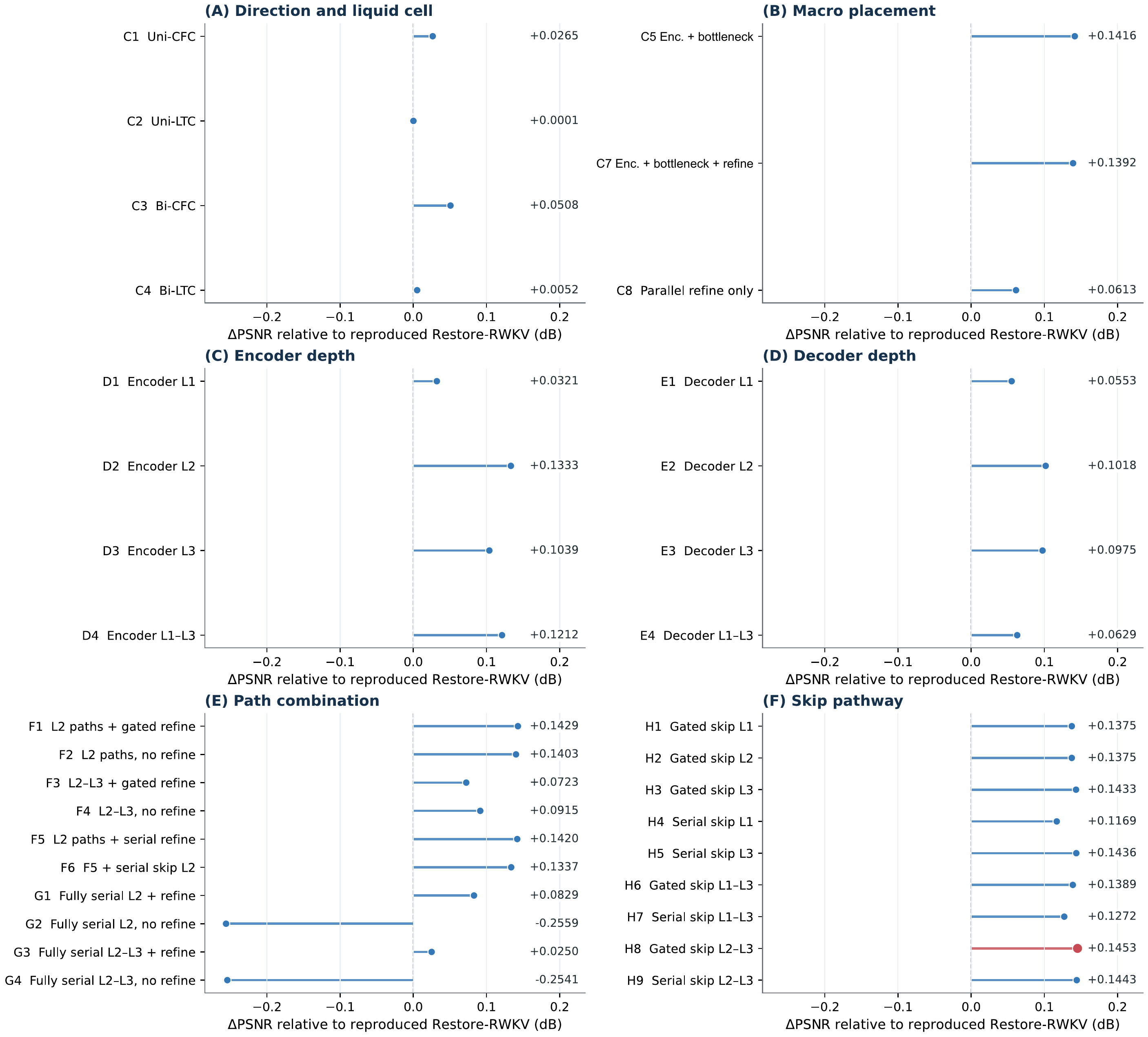}
\caption{CT architecture ablation, expressed as PSNR differences from Restore-RWKV. (A) Propagation direction and cell type. (B) Bottleneck and refinement placement. (C, D) Encoder and decoder depth. (E) Pathway combinations. (F) Skip pathways. H8 denotes the selected topology.}
\label{fig:supp_architecture_screening}
\end{figure}

The architecture study examines propagation direction and recurrent cell (Fig.~\ref{fig:supp_architecture_screening}A), bottleneck and refinement placement (B), encoder and decoder depth (C, D), insertion mode (E), and skip pathways (F). The complete CT results are provided in Table~\ref{tab:supp_full_architecture}. The grouped PSNR differences connect each family of alternatives to its insertion-path decision; H8 is the selected topology.

Intermediate encoder and decoder pathways provide more effective insertion sites than extending liquid modules across all feature levels. Fully serial integration also yields lower fidelity than the selected combination of serial bottleneck processing and gated feature updates. Within the skip-path variants, gated insertion at L2 and L3 gives the highest PSNR and lowest RMSE, supporting the H8 topology.

\small
% Editor: Jialei.He
\setlength{\tabcolsep}{4pt}
\begin{longtable}{p{2.05cm}cp{6.35cm}rrrr}
\caption{CT architecture ablation. Results use the same Restore-RWKV reference and evaluation protocol; $\Delta$PSNR is relative to this reference. The selected configuration is marked in bold.}\label{tab:supp_full_architecture}\\
\toprule
Group & ID & Configuration & PSNR $\uparrow$ & SSIM $\uparrow$ & RMSE $\downarrow$ & $\Delta$PSNR \\
\midrule
\endfirsthead
\multicolumn{7}{l}{\textit{Table~\ref{tab:supp_full_architecture} continued}}\\
\toprule
Group & ID & Configuration & PSNR $\uparrow$ & SSIM $\uparrow$ & RMSE $\downarrow$ & $\Delta$PSNR \\
\midrule
\endhead
\midrule
\multicolumn{7}{r}{\textit{Continued on next page}}\\
\endfoot
\bottomrule
\endlastfoot
Baseline & A-CT & Restore-RWKV, no slice-axis module & 33.7642 & 0.9189 & 8.3950 & 0.0000 \\
\midrule
Direction/core & C1 & One-direction CfC & 33.7907 & 0.9193 & 8.3676 & +0.0265 \\
 & C2 & One-direction LTC & 33.7643 & 0.9187 & 8.3928 & +0.0001 \\
 & C3 & Bidirectional CfC & 33.8150 & 0.9193 & 8.3446 & +0.0508 \\
 & C4 & Bidirectional LTC & 33.7694 & 0.9187 & 8.3889 & +0.0052 \\
\midrule
Bottleneck/\newline refinement & C5 & Strict encoder insertion with bottleneck CfC & 33.9058 & 0.9204 & 8.2586 & +0.1416 \\
 & C7 & Encoder, bottleneck, and gated refinement CfC & 33.9034 & 0.9203 & 8.2606 & +0.1392 \\
 & C8 & Parallel refinement CfC only & 33.8255 & 0.9196 & 8.3355 & +0.0613 \\
\midrule
Encoder depth & D1 & Encoder L1 only & 33.7963 & 0.9187 & 8.3639 & +0.0321 \\
 & D2 & Encoder L2 only & 33.8975 & 0.9203 & 8.2672 & +0.1333 \\
 & D3 & Encoder L3 only & 33.8681 & 0.9199 & 8.2937 & +0.1039 \\
 & D4 & Encoder L1+L2+L3 & 33.8854 & 0.9198 & 8.2777 & +0.1212 \\
\midrule
Decoder depth & E1 & Decoder L1 only & 33.8195 & 0.9195 & 8.3414 & +0.0553 \\
 & E2 & Decoder L2 only & 33.8660 & 0.9200 & 8.2967 & +0.1018 \\
 & E3 & Decoder L3 only & 33.8617 & 0.9199 & 8.3001 & +0.0975 \\
 & E4 & Decoder L1+L2+L3 & 33.8271 & 0.9193 & 8.3330 & +0.0629 \\
\midrule
Path combination & F1 & Encoder L2 + bottleneck + decoder L2 + gated refinement & 33.9071 & 0.9205 & 8.2582 & +0.1429 \\
 & F2 & Encoder L2 + bottleneck + decoder L2, no refinement & 33.9045 & 0.9204 & 8.2604 & +0.1403 \\
 & F3 & Encoder L2/L3 + bottleneck + decoder L2/L3 + gated refinement & 33.8365 & 0.9195 & 8.3242 & +0.0723 \\
 & F4 & Encoder L2/L3 + bottleneck + decoder L2/L3, no refinement & 33.8557 & 0.9199 & 8.3062 & +0.0915 \\
 & F5 & Encoder L2 + bottleneck + decoder L2 + serial refinement & 33.9062 & 0.9205 & 8.2588 & +0.1420 \\
 & F6 & F5 with serial CfC on skip L2 & 33.8979 & 0.9203 & 8.2663 & +0.1337 \\
 & G1 & Fully serial encoder L2 + bottleneck + decoder L2 + refinement & 33.8471 & 0.9200 & 8.3123 & +0.0829 \\
 & G2 & Fully serial encoder L2 + bottleneck + decoder L2, no refinement & 33.5083 & 0.9158 & 8.6396 & -0.2559 \\
 & G3 & Fully serial encoder L2/L3 + bottleneck + decoder L2/L3 + refinement & 33.7892 & 0.9192 & 8.3691 & +0.0250 \\
 & G4 & Fully serial encoder L2/L3 + bottleneck + decoder L2/L3, no refinement & 33.5101 & 0.9157 & 8.6380 & -0.2541 \\
\midrule
Skip pathway & H1 & Gated skip L1 on the F1 backbone & 33.9017 & 0.9205 & 8.2630 & +0.1375 \\
 & H2 & Gated skip L2 on the F1 backbone & 33.9017 & 0.9204 & 8.2631 & +0.1375 \\
 & H3 & Gated skip L3 on the F1 backbone & 33.9075 & 0.9205 & 8.2576 & +0.1433 \\
 & H4 & Serial skip L1 on the F1 backbone & 33.8811 & 0.9201 & 8.2821 & +0.1169 \\
 & H5 & Serial skip L3 on the F1 backbone & 33.9078 & 0.9205 & 8.2577 & +0.1436 \\
 & H6 & Gated skip L1+L2+L3 on the F1 backbone & 33.9031 & 0.9204 & 8.2615 & +0.1389 \\
 & H7 & Serial skip L1+L2+L3 on the F1 backbone & 33.8914 & 0.9203 & 8.2732 & +0.1272 \\
 & H8 & \textbf{Gated skip L2+L3 on the F1 backbone (selected)} & \textbf{33.9095} & 0.9205 & \textbf{8.2560} & \textbf{+0.1453} \\
 & H9 & Serial skip L2+L3 on the F1 backbone & 33.9085 & 0.9204 & 8.2568 & +0.1443 \\
\end{longtable}

\normalsize

\section{Objective decomposition}
\label{sec:supp_objective}
The component-wise CT ablation isolates the contribution of each supervision term under the selected architecture. Each setting in Table~\ref{tab:supp_loss_ablation} removes one term while preserving the remaining model, initialization, and training settings. Removing cross-slice consistency causes the largest PSNR reduction. The second-order term and 2D backbone distillation each provide further improvement, supporting their complementary roles under the selected architecture.

% Editor: Jialei.He
\begin{table}[H]
\centering
\caption{CT loss-component ablation under the selected architecture, seed 42. Components are removed from the full objective one at a time. Bold indicates the best value in each metric column.}
\label{tab:supp_loss_ablation}
\small
\setlength{\tabcolsep}{5pt}
\renewcommand{\arraystretch}{1.08}
\begin{tabular}{lccc}
\toprule
Setting & PSNR $\uparrow$ & SSIM $\uparrow$ & RMSE $\downarrow$ \\
\midrule
Full objective & \textbf{33.954120} & \textbf{0.921013} & \textbf{8.213643} \\
Without distillation & 33.923577 & 0.920509 & 8.241692 \\
Without cross-slice consistency & 33.874285 & 0.919479 & 8.290462 \\
Without second-order consistency & 33.910046 & 0.920077 & 8.256208 \\
\bottomrule
\end{tabular}
\end{table}

\section{Sequence length}
\label{sec:supp_sequence}
\begin{figure}[H]
\centering
\includegraphics[width=0.84\textwidth]{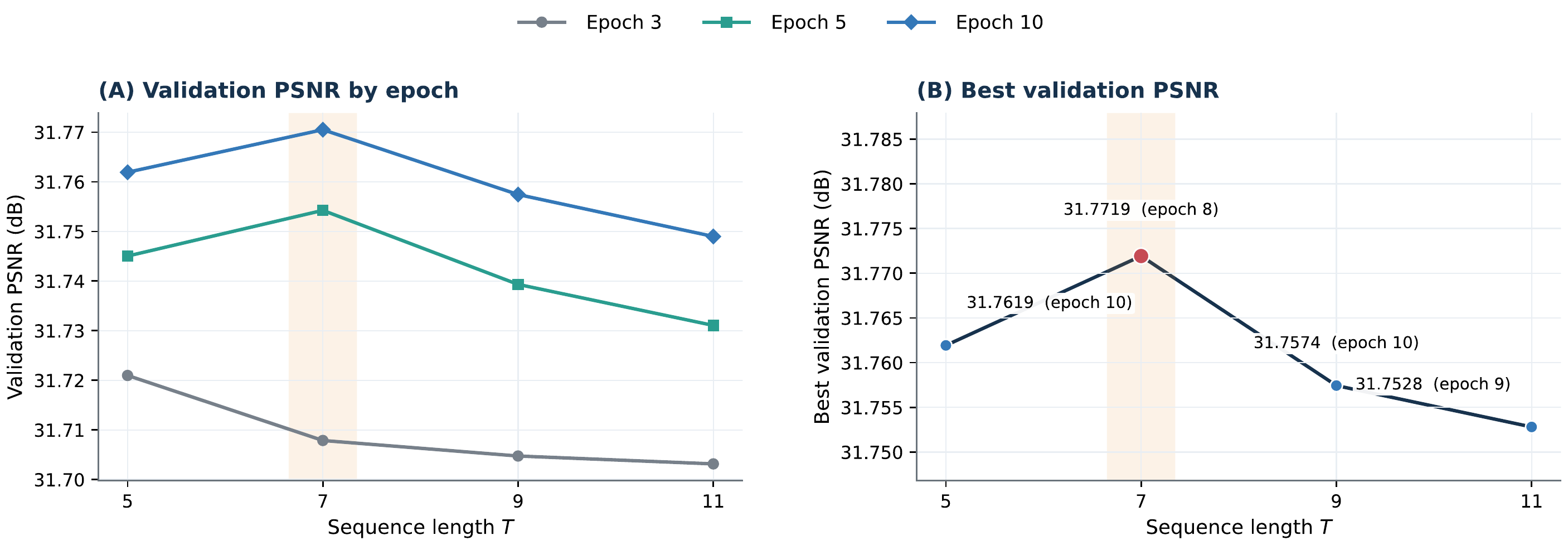}
\caption{CT sequence-length analysis. (A) Validation PSNR at epochs 3, 5, and 10. (B) Highest validation PSNR, with the corresponding epoch annotated. Shading identifies the selected seven-slice window.}
\label{fig:supp_sequence_length}
\end{figure}

The CT sequence-length study evaluates $T\in\{5,7,9,11\}$ using training seed 42 and a 10-epoch budget. Validation PSNR at epochs 3, 5, and 10, the highest validation PSNR, and the test metrics are reported in Table~\ref{tab:supp_sequence_length}. Seven slices achieved the highest validation PSNR. The validation trajectories and their highest observed values are displayed in panels A and B of Fig.~\ref{fig:supp_sequence_length}, respectively.

% Editor: Jialei.He
\begin{table}[H]
\centering
\caption{CT sequence-length analysis with training seed 42 and a 10-epoch budget. Validation and test each use one case. Bold marks the selected seven-slice window.}
\label{tab:supp_sequence_length}
\small
\setlength{\tabcolsep}{4.2pt}
\begin{tabular}{crrrrrrr}
\toprule
$T$ & Val. epoch 3 & Val. epoch 5 & Val. epoch 10 & Best val. (epoch) & Test PSNR & Test SSIM & Test RMSE \\
\midrule
5 & 31.7210 & 31.7450 & 31.7619 & 31.7619 (10) & 33.8825 & 0.9202 & 8.2806 \\
\textbf{7} & \textbf{31.7079} & \textbf{31.7542} & \textbf{31.7705} & \textbf{31.7719 (8)} & \textbf{33.8941} & \textbf{0.9201} & \textbf{8.2697} \\
9 & 31.7047 & 31.7393 & 31.7574 & 31.7574 (10) & 33.8760 & 0.9200 & 8.2865 \\
11 & 31.7032 & 31.7311 & 31.7490 & 31.7528 (9) & 33.8528 & 0.9194 & 8.3079 \\
\bottomrule
\end{tabular}
\end{table}

\section{Context perturbations}
\label{sec:supp_context}
The PET context study evaluates ordered context (default), reversed local order, center repetition and nonlocal same-case sampling using the same trained ContiLNN-RWKV checkpoint. The target center image and the ten evaluated center positions within each of the 29 test scans remain fixed. Table~\ref{tab:supp_context_deltas} reports mean changes relative to ordered context with 95\% case-bootstrap confidence intervals. Table~\ref{tab:supp_context_case_records} provides the 116 case--condition records, including absolute PSNR, SSIM and RMSE and signed differences from the ordered result for the same case. For these signed differences, negative PSNR or SSIM values and positive RMSE values indicate deterioration; the loss convention used in Table~\ref{tab:supp_context_deltas} is stated in its caption.

% Editor: Jialei.He
\begin{table}[H]
\centering
\caption{PET case-level changes relative to the common ordered-context reference. The same trained checkpoint is used in all conditions. Values are means with 95\% percentile confidence intervals from 10,000 case-bootstrap resamples. Positive values denote PSNR or SSIM loss, or RMSE increase. Per-case absolute metrics and signed changes are reported in Table~\ref{tab:supp_context_case_records}.}
\label{tab:supp_context_deltas}
\small
\setlength{\tabcolsep}{4.6pt}
\begin{tabular}{lccc}
\toprule
Perturbation & PSNR loss (dB) & SSIM loss & RMSE increase \\
\midrule
Reversed & 0.2565 [0.2160, 0.2962] & 0.0019 [0.0015, 0.0022] & 0.0022 [0.0017, 0.0027] \\
Center-repeated & 2.6151 [2.4952, 2.7333] & 0.0178 [0.0160, 0.0197] & 0.0229 [0.0211, 0.0246] \\
Nonlocal same-case & 6.3511 [6.0307, 6.6710] & 0.0750 [0.0681, 0.0820] & 0.0770 [0.0725, 0.0816] \\
\bottomrule
\end{tabular}
\end{table}

\section{PET Stage-A trajectory}
\label{sec:supp_stagea}
\begin{figure}[H]
\centering
\includegraphics[width=0.88\textwidth]{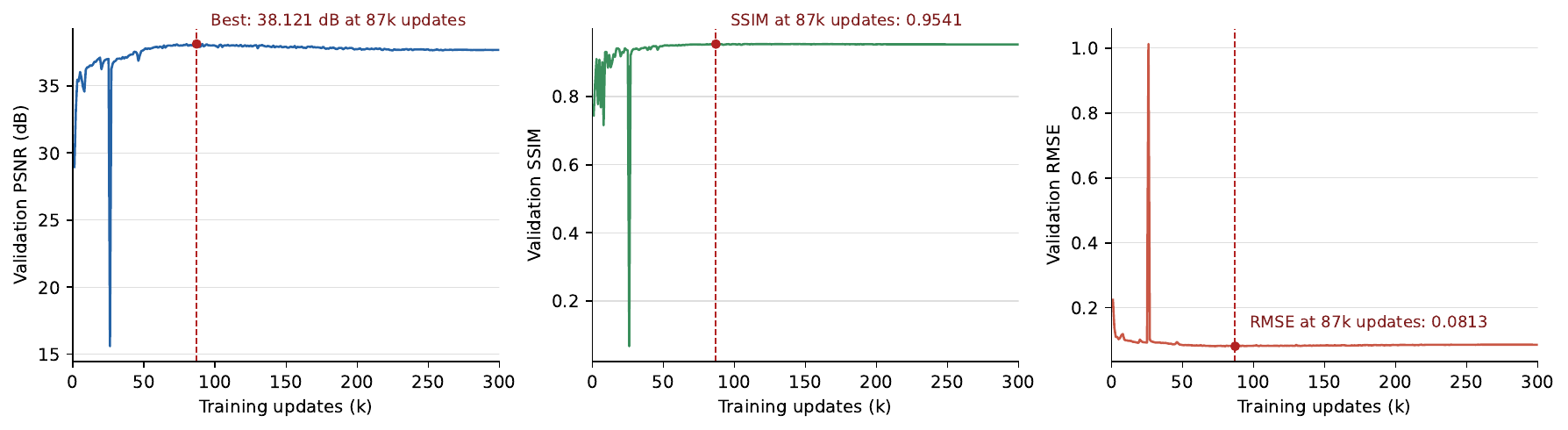}
\caption{PET Stage-A validation trajectory across 300,000 optimizer updates. The dashed line marks the highest validation PSNR at update 87,000. All 300 recorded values are displayed for PSNR, SSIM, and RMSE.}
\label{fig:supp_pet_stagea_convergence}
\end{figure}

The PET Restore-RWKV baseline completed 300,000 optimizer updates, with validation every 1,000 updates. The complete validation trajectory is provided in Fig.~\ref{fig:supp_pet_stagea_convergence}. PSNR reached its highest value of 38.1207 dB at update 87,000, with SSIM 0.9541 and root mean squared error (RMSE) 0.0813. The remaining 213 validation evaluations did not exceed that PSNR, and the checkpoint at update 87,000 was retained for subsequent refinement.

\clearpage
\section{MRI orthogonal views}
\label{sec:supp_mri_orthogonal}

\begin{figure}[H]
\centering
\includegraphics[width=\textwidth]{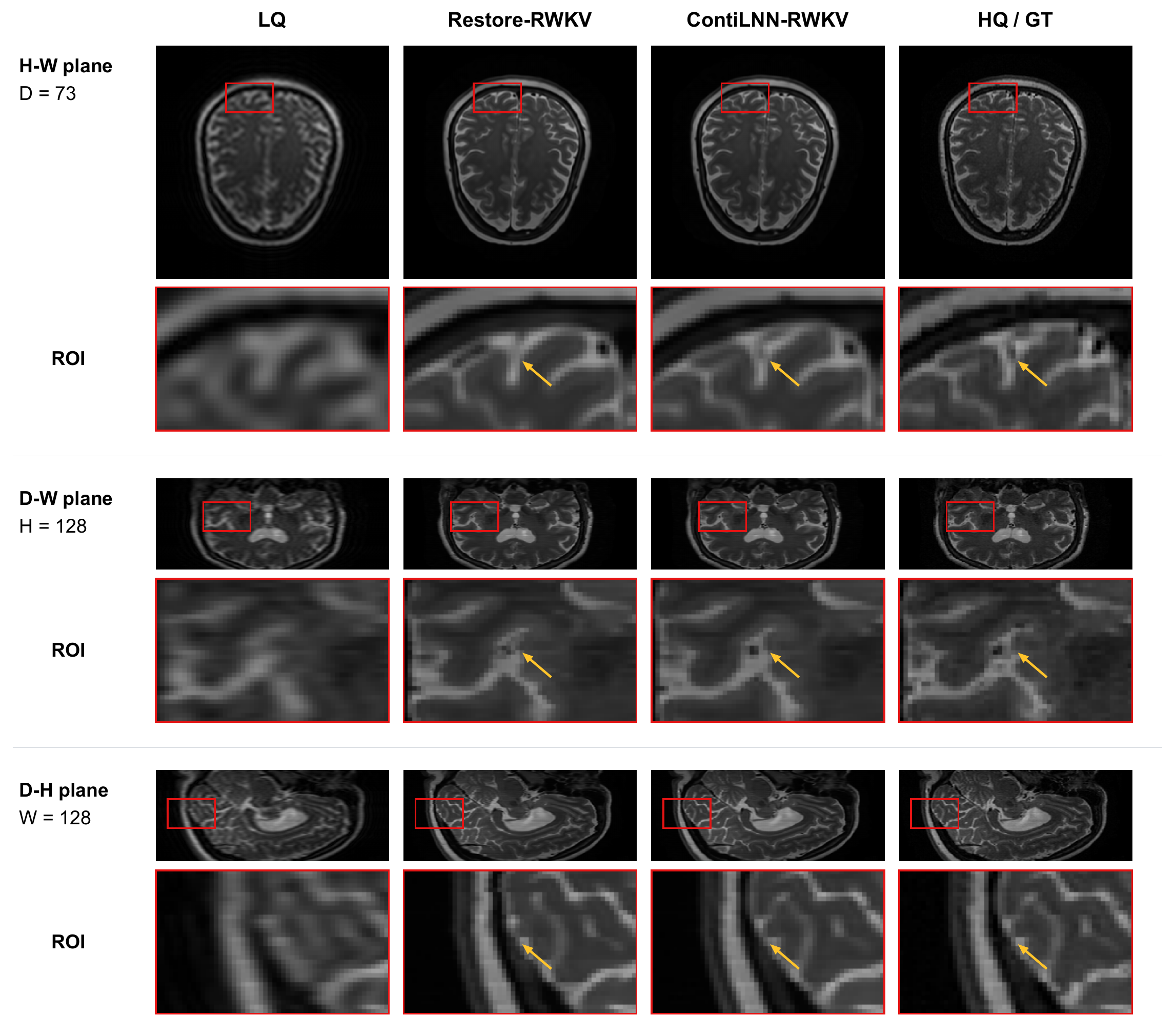}
\caption{MRI restoration in three orthogonal index-space planes for IXI574-IOP-1156-T2. Columns show LQ, Restore-RWKV, ContiLNN-RWKV and HQ images. Red boxes mark matched regions of interest, enlarged below; yellow arrows indicate corresponding local structures. $D$, $H$ and $W$ denote slice index, row and column, respectively (zero-based). The full planes retain the complete available field of view across the 100 processed slices. All panels share the display range $[0,4095]$ and use index-space aspect ratios; physical spacing is unavailable in the processed data.}
\label{fig:supp_mri_ixi574}
\end{figure}

\clearpage
\begin{figure}[H]
\centering
\includegraphics[width=\textwidth]{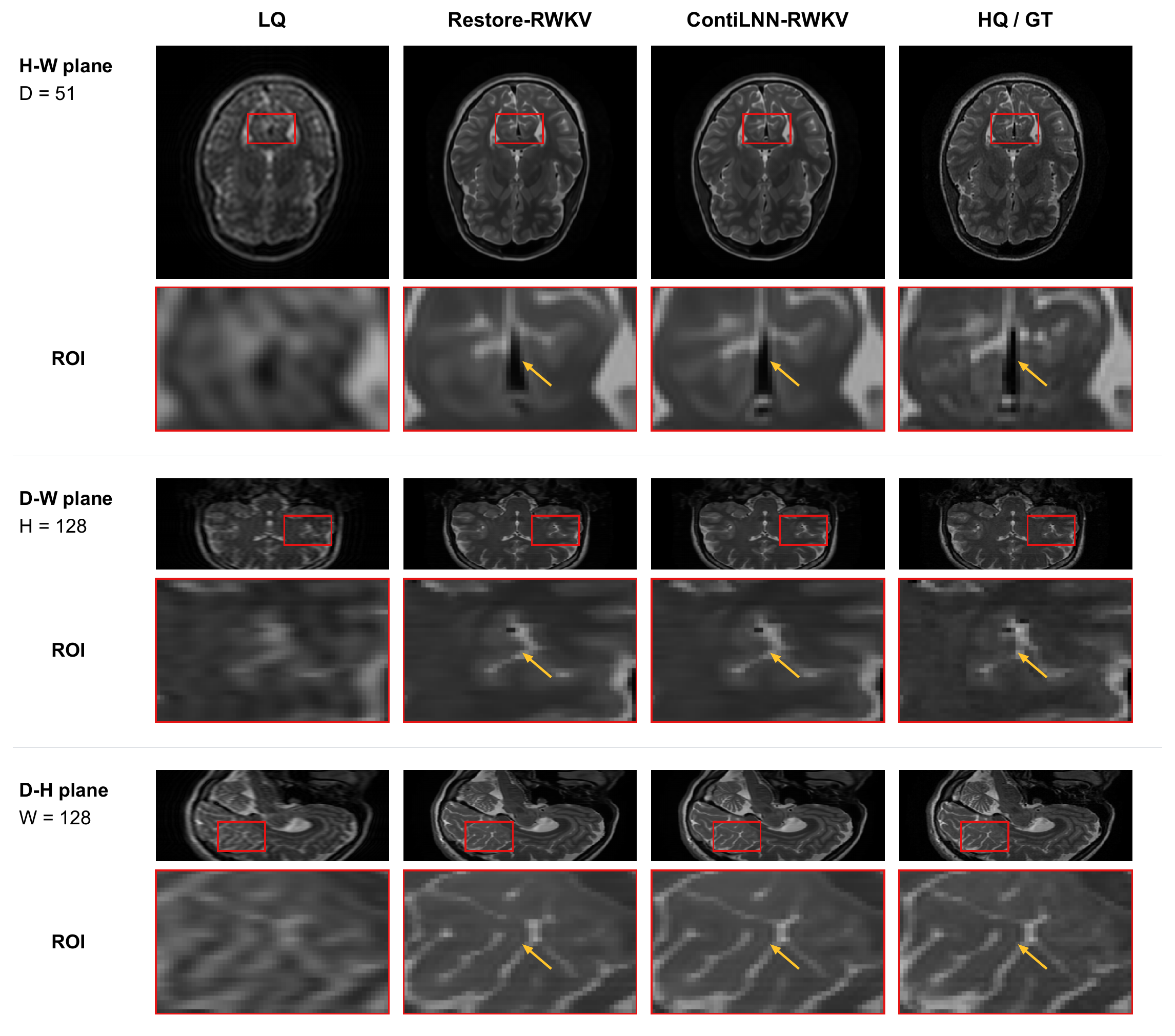}
\caption{MRI restoration in three orthogonal index-space planes for IXI230-IOP-0869-T2. Red-boxed regions are enlarged below each full plane, with yellow arrows marking corresponding local structures. The four columns use identical coordinates and display settings. As in Fig.~\ref{fig:supp_mri_ixi574}, $D$, $H$ and $W$ denote zero-based slice, row and column indices; all 100 processed slices are retained in the $D$-axis reformats. The display range is $[0,4095]$, and aspect ratios follow array indices rather than physical distances.}
\label{fig:supp_mri_ixi230}
\end{figure}

\clearpage
\section{Additional mixed-gap training study}
\label{sec:supp_mixed_gap}
Mixed-gap training examines whether exposure to varied slice-index gaps improves restoration when neighboring context becomes sparse or irregular (Table~\ref{tab:supp_mixed_gap_operators}). Relative to contiguous training, Bi-CfC gains 0.5525, 1.3300 and 0.4032 dB in PSNR at $g=2$, $g=3$ and irregular sampling, respectively. The corresponding gains for Bi-GRU are 0.6530, 1.4685 and 0.4535 dB. Both operators also improve in SSIM and RMSE in these three settings.

At $g=1$, contiguous training retains PSNR advantages of 0.1568 dB for Bi-CfC and 0.0635 dB for Bi-GRU. After mixed-gap training, Bi-GRU achieves marginally higher PSNR at the three uniform gaps, whereas Bi-CfC retains higher SSIM across all four contexts and performs better in all three metrics under irregular sampling.

These results identify mixed-gap training as a complementary strategy for restoration with partially available neighboring context. The benefit extends to both sequence operators, indicating that exposure to varied training gaps contributes to the observed adaptation.

% Editor: Jialei.He
\begin{table}[H]
\centering
\caption{Additional PET sampling experiments on the 29-case test cohort. Both operators are evaluated after contiguous and mixed-gap training at uniform gaps $g=1,2,3$ and irregular combinations of these gaps. Mixed-gap training uses a constant gap within each window and varies the gap across windows. Bold indicates the better training strategy for each operator and test context.}
\label{tab:supp_mixed_gap_operators}
\small
\setlength{\tabcolsep}{7pt}
\renewcommand{\arraystretch}{1.10}
\begin{tabular}{lllrrr}
\toprule
Operator & Test context & Training & PSNR $\uparrow$ & SSIM $\uparrow$ & RMSE $\downarrow$ \\
\midrule
Bi-GRU & $g=1$ & Contiguous & \textbf{38.5176} & \textbf{0.9538} & \textbf{0.0753} \\
 & & Mixed-gap & 38.4541 & 0.9522 & 0.0757 \\
\cmidrule(lr){2-6}
 & $g=2$ & Contiguous & 37.6217 & 0.9462 & 0.0826 \\
 & & Mixed-gap & \textbf{38.2747} & \textbf{0.9517} & \textbf{0.0772} \\
\cmidrule(lr){2-6}
 & $g=3$ & Contiguous & 36.5295 & 0.9366 & 0.0940 \\
 & & Mixed-gap & \textbf{37.9980} & \textbf{0.9504} & \textbf{0.0796} \\
\cmidrule(lr){2-6}
 & Irregular & Contiguous & 37.8544 & 0.9465 & 0.0802 \\
 & & Mixed-gap & \textbf{38.3080} & \textbf{0.9518} & \textbf{0.0768} \\
\midrule
Bi-CfC & $g=1$ & Contiguous & \textbf{38.5868} & \textbf{0.9546} & \textbf{0.0744} \\
 & & Mixed-gap & 38.4301 & 0.9527 & 0.0759 \\
\cmidrule(lr){2-6}
 & $g=2$ & Contiguous & 37.7136 & 0.9476 & 0.0817 \\
 & & Mixed-gap & \textbf{38.2661} & \textbf{0.9521} & \textbf{0.0772} \\
\cmidrule(lr){2-6}
 & $g=3$ & Contiguous & 36.6652 & 0.9393 & 0.0927 \\
 & & Mixed-gap & \textbf{37.9952} & \textbf{0.9510} & \textbf{0.0796} \\
\cmidrule(lr){2-6}
 & Irregular & Contiguous & 37.9494 & 0.9479 & 0.0793 \\
 & & Mixed-gap & \textbf{38.3527} & \textbf{0.9522} & \textbf{0.0764} \\
\bottomrule
\end{tabular}
\end{table}

\clearpage
\section{Contiguous-training performance and inference cost}
\label{sec:supp_contiguous_comparison}

\begin{figure}[H]
\centering
\includegraphics[width=\textwidth]{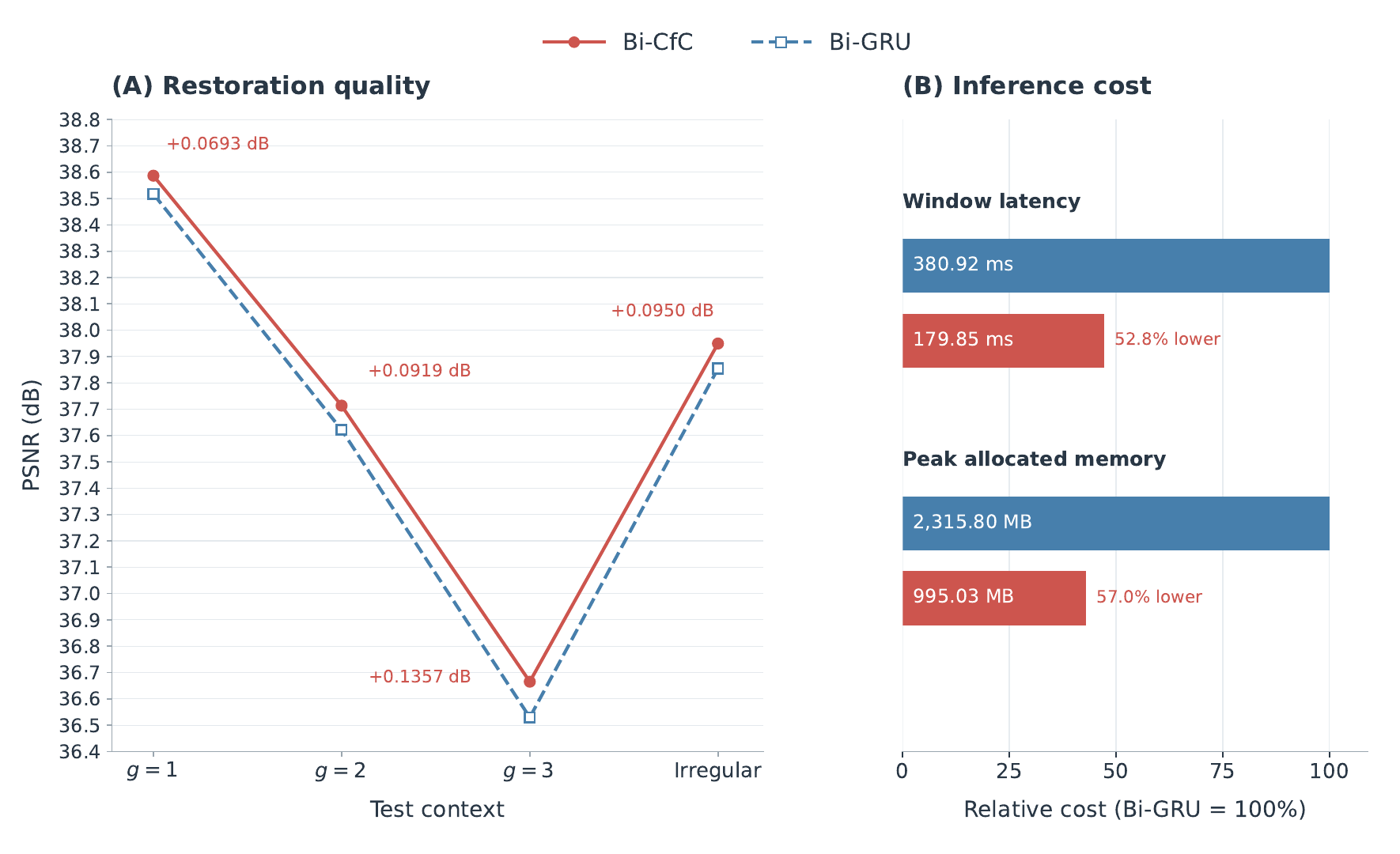}
\caption{Bi-CfC and Bi-GRU trained on contiguous slices. (A) Mean PET PSNR across 29 test cases under four sampling conditions (main Table~2); annotations show Bi-CfC minus Bi-GRU. Lines connect categorical conditions for visual comparison. (B) Seven-slice window latency and peak allocated GPU memory under the main Table~4 profiling conditions, normalized to Bi-GRU; labels report absolute values and relative reductions.}
\label{fig:supp_sampling_cost}
\end{figure}

\clearpage
\section{Case-level PET context results}
\label{sec:supp_context_case_records}

% Editor: Jialei.He
% Case-level values transcribed from the existing Supplementary Data S1.
% This table replaces the separately submitted CSV; no metrics are recomputed.
\begingroup
\small
\setlength{\tabcolsep}{4pt}
\renewcommand{\arraystretch}{1.0}
\setlength{\LTleft}{\fill}
\setlength{\LTright}{\fill}
\setlength{\LTcapwidth}{\textwidth}
\begin{longtable}{llrrrrrr}
\caption{PET context-perturbation results for all 29 test cases under four conditions (116 records). Each metric is the mean over the same ten evaluated center slices within a case. $\Delta$ denotes the condition result minus the ordered-context result for the same case; negative PSNR or SSIM changes and positive RMSE changes indicate deterioration. PSNR and its change are in dB; SSIM is dimensionless and RMSE retains the original evaluation units. Metrics and differences are rounded independently to six decimal places.}
\label{tab:supp_context_case_records}\\
\toprule
Case & Context & PSNR & $\Delta$PSNR & SSIM & $\Delta$SSIM & RMSE & $\Delta$RMSE \\
\midrule
\endfirsthead
\multicolumn{8}{l}{\textbf{Table \thetable\ continued}}\\
\toprule
Case & Context & PSNR & $\Delta$PSNR & SSIM & $\Delta$SSIM & RMSE & $\Delta$RMSE \\
\midrule
\endhead
\midrule
\multicolumn{8}{r}{\footnotesize Continued on the next page}\\
\endfoot
\bottomrule
\endlastfoot
Patient 10 & Ordered (default) & 36.761593 & 0.000000 & 0.948692 & 0.000000 & 0.073840 & 0.000000 \\*
Patient 10 & Reversed & 36.449061 & -0.312532 & 0.946832 & -0.001860 & 0.077593 & 0.003753 \\*
Patient 10 & Center-repeated & 34.391667 & -2.369926 & 0.928067 & -0.020624 & 0.094301 & 0.020461 \\*
Patient 10 & Nonlocal same-case & 31.008852 & -5.752742 & 0.859689 & -0.089003 & 0.142358 & 0.068518 \\
\addlinespace[1pt]
Patient 101 & Ordered (default) & 38.906556 & 0.000000 & 0.952976 & 0.000000 & 0.062351 & 0.000000 \\*
Patient 101 & Reversed & 38.850965 & -0.055590 & 0.952068 & -0.000908 & 0.062559 & 0.000207 \\*
Patient 101 & Center-repeated & 36.193087 & -2.713469 & 0.935912 & -0.017064 & 0.081845 & 0.019493 \\*
Patient 101 & Nonlocal same-case & 33.117327 & -5.789229 & 0.888393 & -0.064584 & 0.115678 & 0.053326 \\
\addlinespace[1pt]
Patient 103 & Ordered (default) & 39.423093 & 0.000000 & 0.958721 & 0.000000 & 0.061477 & 0.000000 \\*
Patient 103 & Reversed & 39.186257 & -0.236837 & 0.957317 & -0.001405 & 0.063164 & 0.001687 \\*
Patient 103 & Center-repeated & 36.511077 & -2.912017 & 0.941561 & -0.017161 & 0.083987 & 0.022510 \\*
Patient 103 & Nonlocal same-case & 31.880395 & -7.542698 & 0.887479 & -0.071242 & 0.149856 & 0.088379 \\
\addlinespace[1pt]
Patient 106 & Ordered (default) & 38.440922 & 0.000000 & 0.957367 & 0.000000 & 0.084605 & 0.000000 \\*
Patient 106 & Reversed & 38.069264 & -0.371658 & 0.954919 & -0.002448 & 0.087901 & 0.003296 \\*
Patient 106 & Center-repeated & 35.742716 & -2.698206 & 0.939603 & -0.017764 & 0.106971 & 0.022366 \\*
Patient 106 & Nonlocal same-case & 31.840640 & -6.600282 & 0.877543 & -0.079823 & 0.168198 & 0.083593 \\
\addlinespace[1pt]
Patient 112 & Ordered (default) & 37.647933 & 0.000000 & 0.949370 & 0.000000 & 0.079394 & 0.000000 \\*
Patient 112 & Reversed & 37.213270 & -0.434663 & 0.947025 & -0.002346 & 0.084175 & 0.004781 \\*
Patient 112 & Center-repeated & 34.939279 & -2.708655 & 0.930451 & -0.018919 & 0.104146 & 0.024752 \\*
Patient 112 & Nonlocal same-case & 30.240319 & -7.407614 & 0.845936 & -0.103434 & 0.174975 & 0.095581 \\
\addlinespace[1pt]
Patient 124 & Ordered (default) & 35.552508 & 0.000000 & 0.930709 & 0.000000 & 0.082582 & 0.000000 \\*
Patient 124 & Reversed & 35.216282 & -0.336226 & 0.927213 & -0.003497 & 0.085478 & 0.002896 \\*
Patient 124 & Center-repeated & 33.083976 & -2.468532 & 0.907105 & -0.023605 & 0.105432 & 0.022850 \\*
Patient 124 & Nonlocal same-case & 29.728979 & -5.823529 & 0.834091 & -0.096619 & 0.162336 & 0.079753 \\
\addlinespace[1pt]
Patient 125 & Ordered (default) & 34.246244 & 0.000000 & 0.920016 & 0.000000 & 0.051651 & 0.000000 \\*
Patient 125 & Reversed & 33.987126 & -0.259118 & 0.916571 & -0.003445 & 0.053259 & 0.001608 \\*
Patient 125 & Center-repeated & 32.412747 & -1.833497 & 0.903842 & -0.016174 & 0.062665 & 0.011014 \\*
Patient 125 & Nonlocal same-case & 29.153594 & -5.092650 & 0.833906 & -0.086110 & 0.092980 & 0.041330 \\
\addlinespace[1pt]
Patient 126 & Ordered (default) & 34.311084 & 0.000000 & 0.927582 & 0.000000 & 0.096870 & 0.000000 \\*
Patient 126 & Reversed & 34.131184 & -0.179900 & 0.924329 & -0.003253 & 0.099125 & 0.002254 \\*
Patient 126 & Center-repeated & 32.056365 & -2.254719 & 0.899263 & -0.028319 & 0.123565 & 0.026695 \\*
Patient 126 & Nonlocal same-case & 29.613399 & -4.697686 & 0.810051 & -0.117531 & 0.166570 & 0.069699 \\
\addlinespace[1pt]
Patient 129 & Ordered (default) & 35.778501 & 0.000000 & 0.937247 & 0.000000 & 0.079079 & 0.000000 \\*
Patient 129 & Reversed & 35.400782 & -0.377718 & 0.934703 & -0.002544 & 0.083189 & 0.004110 \\*
Patient 129 & Center-repeated & 33.405693 & -2.372808 & 0.916115 & -0.021132 & 0.100654 & 0.021576 \\*
Patient 129 & Nonlocal same-case & 30.415584 & -5.362917 & 0.836156 & -0.101091 & 0.146956 & 0.067877 \\
\addlinespace[1pt]
Patient 142 & Ordered (default) & 35.583863 & 0.000000 & 0.937662 & 0.000000 & 0.077916 & 0.000000 \\*
Patient 142 & Reversed & 35.191704 & -0.392158 & 0.934590 & -0.003072 & 0.081630 & 0.003714 \\*
Patient 142 & Center-repeated & 33.089205 & -2.494658 & 0.913623 & -0.024039 & 0.100648 & 0.022732 \\*
Patient 142 & Nonlocal same-case & 30.329091 & -5.254772 & 0.847152 & -0.090510 & 0.143359 & 0.065443 \\
\addlinespace[1pt]
Patient 145 & Ordered (default) & 38.446787 & 0.000000 & 0.959709 & 0.000000 & 0.075416 & 0.000000 \\*
Patient 145 & Reversed & 38.160992 & -0.285794 & 0.957976 & -0.001733 & 0.078248 & 0.002832 \\*
Patient 145 & Center-repeated & 35.758012 & -2.688774 & 0.942517 & -0.017193 & 0.099953 & 0.024537 \\*
Patient 145 & Nonlocal same-case & 31.776824 & -6.669962 & 0.867212 & -0.092497 & 0.158932 & 0.083515 \\
\addlinespace[1pt]
Patient 146 & Ordered (default) & 37.617776 & 0.000000 & 0.951471 & 0.000000 & 0.098457 & 0.000000 \\*
Patient 146 & Reversed & 37.203507 & -0.414269 & 0.949493 & -0.001978 & 0.103322 & 0.004865 \\*
Patient 146 & Center-repeated & 35.446814 & -2.170962 & 0.934857 & -0.016614 & 0.121174 & 0.022717 \\*
Patient 146 & Nonlocal same-case & 32.671686 & -4.946090 & 0.888647 & -0.062825 & 0.165737 & 0.067280 \\
\addlinespace[1pt]
Patient 15 & Ordered (default) & 40.962795 & 0.000000 & 0.971458 & 0.000000 & 0.075801 & 0.000000 \\*
Patient 15 & Reversed & 40.712937 & -0.249859 & 0.969814 & -0.001644 & 0.078246 & 0.002445 \\*
Patient 15 & Center-repeated & 38.253667 & -2.709128 & 0.956811 & -0.014647 & 0.103506 & 0.027705 \\*
Patient 15 & Nonlocal same-case & 34.631902 & -6.330893 & 0.923857 & -0.047601 & 0.158798 & 0.082997 \\
\addlinespace[1pt]
Patient 150 & Ordered (default) & 38.129044 & 0.000000 & 0.955725 & 0.000000 & 0.077390 & 0.000000 \\*
Patient 150 & Reversed & 37.880455 & -0.248589 & 0.953694 & -0.002030 & 0.078965 & 0.001575 \\*
Patient 150 & Center-repeated & 35.868792 & -2.260253 & 0.936367 & -0.019358 & 0.095018 & 0.017628 \\*
Patient 150 & Nonlocal same-case & 32.463270 & -5.665774 & 0.872491 & -0.083234 & 0.141190 & 0.063800 \\
\addlinespace[1pt]
Patient 158 & Ordered (default) & 35.070730 & 0.000000 & 0.931075 & 0.000000 & 0.103789 & 0.000000 \\*
Patient 158 & Reversed & 34.668173 & -0.402558 & 0.926429 & -0.004646 & 0.107839 & 0.004050 \\*
Patient 158 & Center-repeated & 32.009709 & -3.061022 & 0.899749 & -0.031326 & 0.140108 & 0.036318 \\*
Patient 158 & Nonlocal same-case & 29.852387 & -5.218343 & 0.833674 & -0.097401 & 0.188120 & 0.084331 \\
\addlinespace[1pt]
Patient 20 & Ordered (default) & 39.811874 & 0.000000 & 0.971265 & 0.000000 & 0.092518 & 0.000000 \\*
Patient 20 & Reversed & 39.621685 & -0.190189 & 0.970323 & -0.000942 & 0.093760 & 0.001242 \\*
Patient 20 & Center-repeated & 37.181179 & -2.630695 & 0.960268 & -0.010997 & 0.122151 & 0.029633 \\*
Patient 20 & Nonlocal same-case & 33.928895 & -5.882979 & 0.924185 & -0.047080 & 0.176277 & 0.083759 \\
\addlinespace[1pt]
Patient 3 & Ordered (default) & 41.752468 & 0.000000 & 0.977102 & 0.000000 & 0.064732 & 0.000000 \\*
Patient 3 & Reversed & 41.638737 & -0.113731 & 0.976417 & -0.000685 & 0.065159 & 0.000428 \\*
Patient 3 & Center-repeated & 39.191530 & -2.560938 & 0.966135 & -0.010968 & 0.086218 & 0.021486 \\*
Patient 3 & Nonlocal same-case & 34.790951 & -6.961518 & 0.914884 & -0.062218 & 0.142699 & 0.077967 \\
\addlinespace[1pt]
Patient 34 & Ordered (default) & 42.556716 & 0.000000 & 0.978345 & 0.000000 & 0.070179 & 0.000000 \\*
Patient 34 & Reversed & 42.264116 & -0.292600 & 0.977177 & -0.001167 & 0.072261 & 0.002082 \\*
Patient 34 & Center-repeated & 39.723645 & -2.833071 & 0.970473 & -0.007872 & 0.096387 & 0.026208 \\*
Patient 34 & Nonlocal same-case & 35.095562 & -7.461154 & 0.928046 & -0.050299 & 0.166897 & 0.096718 \\
\addlinespace[1pt]
Patient 36 & Ordered (default) & 39.139292 & 0.000000 & 0.949931 & 0.000000 & 0.068756 & 0.000000 \\*
Patient 36 & Reversed & 39.024307 & -0.114985 & 0.949092 & -0.000839 & 0.069624 & 0.000868 \\*
Patient 36 & Center-repeated & 37.091457 & -2.047835 & 0.937786 & -0.012145 & 0.085815 & 0.017059 \\*
Patient 36 & Nonlocal same-case & 31.942837 & -7.196455 & 0.884026 & -0.065905 & 0.158198 & 0.089442 \\
\addlinespace[1pt]
Patient 41 & Ordered (default) & 38.679674 & 0.000000 & 0.960796 & 0.000000 & 0.064076 & 0.000000 \\*
Patient 41 & Reversed & 38.537203 & -0.142472 & 0.959681 & -0.001115 & 0.065132 & 0.001057 \\*
Patient 41 & Center-repeated & 36.102895 & -2.576779 & 0.944291 & -0.016505 & 0.083653 & 0.019577 \\*
Patient 41 & Nonlocal same-case & 32.834068 & -5.845606 & 0.900435 & -0.060361 & 0.127958 & 0.063882 \\
\addlinespace[1pt]
Patient 65 & Ordered (default) & 41.390868 & 0.000000 & 0.969749 & 0.000000 & 0.070523 & 0.000000 \\*
Patient 65 & Reversed & 41.104567 & -0.286300 & 0.968606 & -0.001143 & 0.072962 & 0.002439 \\*
Patient 65 & Center-repeated & 38.271384 & -3.119483 & 0.953104 & -0.016645 & 0.097286 & 0.026763 \\*
Patient 65 & Nonlocal same-case & 33.550288 & -7.840580 & 0.893146 & -0.076603 & 0.171536 & 0.101013 \\
\addlinespace[1pt]
Patient 68 & Ordered (default) & 43.532892 & 0.000000 & 0.976940 & 0.000000 & 0.071733 & 0.000000 \\*
Patient 68 & Reversed & 43.363515 & -0.169377 & 0.976288 & -0.000652 & 0.073002 & 0.001269 \\*
Patient 68 & Center-repeated & 40.414518 & -3.118374 & 0.964177 & -0.012763 & 0.099909 & 0.028176 \\*
Patient 68 & Nonlocal same-case & 36.863854 & -6.669038 & 0.926691 & -0.050249 & 0.153561 & 0.081828 \\
\addlinespace[1pt]
Patient 78 & Ordered (default) & 40.748344 & 0.000000 & 0.965842 & 0.000000 & 0.077276 & 0.000000 \\*
Patient 78 & Reversed & 40.389416 & -0.358928 & 0.963137 & -0.002705 & 0.080435 & 0.003159 \\*
Patient 78 & Center-repeated & 37.941662 & -2.806682 & 0.943516 & -0.022327 & 0.105228 & 0.027951 \\*
Patient 78 & Nonlocal same-case & 34.737163 & -6.011181 & 0.919173 & -0.046669 & 0.159874 & 0.082597 \\
\addlinespace[1pt]
Patient 88 & Ordered (default) & 37.900479 & 0.000000 & 0.945639 & 0.000000 & 0.067376 & 0.000000 \\*
Patient 88 & Reversed & 37.785339 & -0.115139 & 0.944122 & -0.001518 & 0.067898 & 0.000522 \\*
Patient 88 & Center-repeated & 35.205895 & -2.694584 & 0.924304 & -0.021335 & 0.087426 & 0.020050 \\*
Patient 88 & Nonlocal same-case & 31.105505 & -6.794974 & 0.871470 & -0.074169 & 0.147157 & 0.079781 \\
\addlinespace[1pt]
Patient 90 & Ordered (default) & 40.654854 & 0.000000 & 0.955710 & 0.000000 & 0.061844 & 0.000000 \\*
Patient 90 & Reversed & 40.265983 & -0.388872 & 0.953713 & -0.001997 & 0.064289 & 0.002445 \\*
Patient 90 & Center-repeated & 37.573345 & -3.081509 & 0.939819 & -0.015891 & 0.083660 & 0.021817 \\*
Patient 90 & Nonlocal same-case & 33.090787 & -7.564067 & 0.880834 & -0.074876 & 0.138461 & 0.076617 \\
\addlinespace[1pt]
Patient 94 & Ordered (default) & 41.247117 & 0.000000 & 0.962865 & 0.000000 & 0.064122 & 0.000000 \\*
Patient 94 & Reversed & 40.966248 & -0.280869 & 0.961544 & -0.001321 & 0.066236 & 0.002114 \\*
Patient 94 & Center-repeated & 38.533897 & -2.713220 & 0.950953 & -0.011912 & 0.085782 & 0.021661 \\*
Patient 94 & Nonlocal same-case & 34.509461 & -6.737656 & 0.906964 & -0.055901 & 0.142272 & 0.078150 \\
\addlinespace[1pt]
Patient 95 & Ordered (default) & 38.715439 & 0.000000 & 0.953354 & 0.000000 & 0.068347 & 0.000000 \\*
Patient 95 & Reversed & 38.610673 & -0.104766 & 0.952163 & -0.001192 & 0.069059 & 0.000712 \\*
Patient 95 & Center-repeated & 35.812634 & -2.902805 & 0.932745 & -0.020610 & 0.092550 & 0.024203 \\*
Patient 95 & Nonlocal same-case & 31.746436 & -6.969003 & 0.870358 & -0.082997 & 0.148146 & 0.079799 \\
\addlinespace[1pt]
Patient 96 & Ordered (default) & 39.941632 & 0.000000 & 0.965331 & 0.000000 & 0.059480 & 0.000000 \\*
Patient 96 & Reversed & 39.860771 & -0.080861 & 0.964630 & -0.000701 & 0.059839 & 0.000360 \\*
Patient 96 & Center-repeated & 37.788493 & -2.153139 & 0.951008 & -0.014322 & 0.073622 & 0.014142 \\*
Patient 96 & Nonlocal same-case & 32.825961 & -7.115670 & 0.891644 & -0.073686 & 0.132477 & 0.072998 \\
\addlinespace[1pt]
Patient 98 & Ordered (default) & 39.924393 & 0.000000 & 0.956135 & 0.000000 & 0.063137 & 0.000000 \\*
Patient 98 & Reversed & 39.682137 & -0.242256 & 0.954499 & -0.001636 & 0.064639 & 0.001502 \\*
Patient 98 & Center-repeated & 37.042061 & -2.882332 & 0.938419 & -0.017716 & 0.083982 & 0.020845 \\*
Patient 98 & Nonlocal same-case & 32.948196 & -6.976197 & 0.886951 & -0.069183 & 0.137132 & 0.073995 \\
\end{longtable}
\endgroup